\documentclass[12pt,reqno]{amsart}
\pdfoutput=1 

\usepackage{ucs}              
\usepackage[utf8x]{inputenc}    

\usepackage[T1]{fontenc}     
\usepackage{lmodern}         

\usepackage[text={160mm,250mm},centering]{geometry}
\usepackage{url}
\DeclareUrlCommand\email{\urlstyle{tt}}
\usepackage[breaklinks,colorlinks=true,plainpages=false,linkcolor=blue]{hyperref}

\usepackage{latexsym}
\usepackage{amsmath,amssymb,amscd,amsbsy}
\usepackage{upgreek}

\usepackage[british]{babel} 
\usepackage{rcs}  \RCS $Revision: 2.1 $ \RCS $Date: 2025/02/13 19:28:46 $ \RCS $Author: hgm $ \RCS $RCSfile: arxiv-ROM-CEX.tex,v $ \RCS $Id: arxiv-ROM-CEX.tex,v 2.1 2025/02/13 19:28:46 hgm Exp $

\usepackage{tensor}

\usepackage{ifontdef}

\usepackage{amsmath}

\newcommand{\ignore}[1]{}

\newcommand{\Lp}{\mrm{L}}

\newcommand{\vepsilon}{\varepsilon}

\newcommand{\vsigma}{\varsigma}

\newcommand{\vphi}{\varphi}

\newcommand{\vkappa}{\varkappa}

\newcommand{\vek}[1]{\mathchoice{\displaystyle\boldsymbol{#1}}
{\textstyle\boldsymbol{#1}}{\scriptstyle\boldsymbol{#1}}
{\scriptscriptstyle\boldsymbol{#1}}}

\newcommand{\ops}[1]{\mathchoice{\displaystyle\mathsf{#1}}
{\textstyle\mathsf{#1}}{\scriptstyle\mathsf{#1}}
{\scriptscriptstyle\mathsf{#1}}}

\newcommand{\tnb}[1]{\mathchoice{\displaystyle\mathboldsans{#1}}
{\textstyle\mathboldsans{#1}}{\scriptstyle\mathboldsans{#1}}
{\scriptscriptstyle\mathboldsans{#1}}}

\newcommand{\vbar}[1]{\vek{\bar{#1}}}

\newcommand{\vhat}[1]{\vek{\hat{#1}}}

\DeclareMathOperator{\diag}{diag}

\DeclareMathOperator{\im}{im}

\DeclareMathOperator{\rank}{rank}
\DeclareMathOperator{\tr}{tr}

\DeclareMathOperator{\spn}{span}

\DeclareMathOperator{\cl}{cl}

\newcommand{\dd}{\mathop{}\!\partial}
\newcommand{\di}{\mathop{}\!\mathrm{d}}

\newcommand{\EXP}[1]{\mathbb{E}\left(#1\right)}

\newcommand{\ip}[2]{\left\langle #1 , #2 \right\rangle}
\newcommand{\bkt}[2]{\left\langle #1 \mid #2 \right\rangle}

\newcommand{\nd}[1]{\left\Vert #1 \right\Vert}

\newcommand{\trpos}{{\ops{T}}}

\newcommand{\BIGOP}[1]{\mathop{\mathchoice%
{\raise-0.22em\hbox{\huge $#1$}} {\raise-0.05em\hbox{\Large $#1$}}
{\hbox{\large $#1$}}{#1}}}
\newcommand{\BIGboxplus}{\mathop{\mathchoice%
{\raise-0.35em\hbox{\huge $\boxplus$}}%
{\raise-0.15em\hbox{\Large $\boxplus$}}{\hbox{\large
$\boxplus$}}{\boxplus}}}

\newcommand{\bbbone}{{\mathchoice {\rm 1\mskip-4mu l} {\rm 1\mskip-4mu l} {\rm 1\mskip-4.5mu l} {\rm 1\mskip-5mu l}}}

\newcommand{\frkt}[2]{{{\raise0.6ex\hbox{{\leavevmode$\textstyle #1$}}}{\raise0.25ex\hbox{\kern-0.35ex\hbox{/}}}{\raise-0.3ex\hbox{\kern-0.4ex\hbox{{\leavevmode$\textstyle #2$}}}}}}
\newcommand{\frks}[2]{{{\raise0.7ex\hbox{{\leavevmode$\scriptstyle #1$}}}{\raise0.2ex\hbox{\kern-0.4ex\hbox{\footnotesize /}}}{\raise-0.2ex\hbox{\kern-0.4ex\hbox{{\leavevmode$\scriptstyle #2$}}}}}}
\newcommand{\frkx}[2]{{{\raise0.75ex\hbox{{\leavevmode$\scriptscriptstyle #1$}}}{\raise0.17ex\hbox{\kern-0.45ex\hbox{\scriptsize /}}}{\raise-0.15ex\hbox{\kern-0.4ex\hbox{{\leavevmode$\scriptscriptstyle #2$}}}}}}
\newcommand{\frkz}[2]{{{\raise0.75ex\hbox{{\leavevmode$\scriptscriptstyle #1$}}}{\raise0.17ex\hbox{\kern-0.45ex\hbox{\tiny /}}}{\raise-0.15ex\hbox{\kern-0.4ex\hbox{{\leavevmode$\scriptscriptstyle #2$}}}}}}

\newcommand{\frk}[2]{{\mathchoice{{\frkt{#1}{#2}}}{{\frks{#1}{#2}}}{{\frkx{#1}{#2}}}{{\frkz{#1}{#2}}}}}

\ignore{     

\usepackage[sfdefault=cmbr,OMLmathsans]{isomath}  

\newcommand{\ignore}[1]{}

\newcommand{\vek}[1]{\mathchoice{\displaystyle\boldsymbol#1}
{\textstyle\boldsymbol#1}{\scriptstyle\boldsymbol#1}
{\scriptscriptstyle\boldsymbol#1}}

\newcommand{\F}{\mathfrak} 
\newcommand{\C}{\mathcal}  
\newcommand{\mrm}{\mathrm}     
\newcommand{\dd}{\mathop{}\!\partial}
\newcommand{\di}{\mathop{}\!\mathrm{d}}

\newcommand{\ip}[2]{\langle #1 , #2 \rangle}

\newcommand{\bkt}[2]{\langle #1 | #2 \rangle}

\newcommand{\nd}[1]{\| #1 \|}

\DeclareMathOperator{\tr}{tr}
\DeclareMathOperator{\diag}{diag}

}    

\newcommand{\feq}[1]{Eq.(\ref{#1})} 
\newcommand{\feqs}[2]{Eqs.(\ref{#1}) and (\ref{#2})} 
\newcommand{\feqss}[3]{Eqs.(\ref{#1}),(\ref{#2}) and (\ref{#3})} 
\newcommand{\fsec}[1]{Section~\ref{#1}}
\newcommand{\fsecs}[2]{Sections~\ref{#1} and \ref{#2}} 

\newcommand{\feX}[1]{Example~\ref{#1}}

\newcommand{\KL}{Karhunen-Lo\`eve}

\DeclareMathOperator{\var}{var}

\newcommand{\Ex}{\D{E}}

\newcommand{\RR}{\D{R}}

\newcommand{\sigalg}{\F{F}}
\newcommand{\prob}{\D{P}}

\newcommand{\vA}{\vek{A}}

\newcommand{\vC}{\vek{C}}

\newcommand{\vG}{\vek{G}}

\newcommand{\vK}{\vek{K}}

\newcommand{\vR}{\vek{R}}

\newcommand{\vV}{\vek{V}}
\newcommand{\vW}{\vek{W}}
\newcommand{\vX}{\vek{X}}

\newcommand{\vZ}{\vek{Z}}

\newcommand{\tK}{\tnb{K}}

\newcommand{\ve}{\vek{e}}

\newcommand{\vg}{\vek{g}}

\newcommand{\vr}{\vek{r}}

\newcommand{\vv}{\vek{v}}
\newcommand{\vw}{\vek{w}}

\newcommand{\vz}{\vek{z}}

\newcommand{\vmu}{\vek{\mu}}

\newcommand{\Mlg}{\C{M}}
\newcommand{\Uvk}{\C{U}}
\newcommand{\Vvk}{\C{V}}
\newcommand{\Pst}{\C{P}}

\newcommand{\Svk}{\C{S}}

\newcommand{\citem}[1]{\cite{#1}}

\newtheorem{xmpn}{Example}

\begin{document}

\title[Analysing Parametric ROMs and Conditional Expectation]{Reduced Order Models and 
     Conditional Expectation  \\
     --- Analysing Parametric Low-Order Approximations ---}
\author[H.G. Matthies]{Hermann G. Matthies}
\address{Institute of Scientific Computing, Technische Universität Braunschweig, Germany
ORCID: \orcidauthorA, Email: \ttt{H.Matthies@tu-bs.de}}

\date{\today}         

\newcommand{\orcidauthorA}{0000-0002-8644-5574} 



\begin{abstract}
Systems may depend on parameters which one may control, or which serve 
to optimise the system, or are imposed externally,
or they could be uncertain.  This last case 
is taken as the ``Leitmotiv'' for the following.  A reduced order model is 
produced from the full order model by some kind of projection onto a relatively
low-dimensional manifold or subspace.  The parameter 
dependent reduction process produces a function of the parameters into the 
manifold.  One now wants to examine the relation between the full and the reduced state
for all possible parameter values of interest.
Similarly, in the field of machine learning, also a function of the parameter set into the 
image space of the machine learning model is learned  
on a training set of samples, typically minimising the mean-square 
error.  This set may be seen as a sample from some probability distribution, 
and thus the training is an approximate computation of the expectation, giving 
an approximation to the conditional expectation, 
a special case of an Bayesian updating where the Bayesian loss 
function is the mean-square error.  This offers the possibility of having a combined 
look at these methods, and also of introducing more general loss functions.
%
%
\end{abstract}

\maketitle

\section{Introduction}  \label{S:intro}
Parameter dependent systems occur often in numerical simulations.  These parameters may
have different significance: they may be parameters which one may have to determine to
tune the simulation, they may be used for control or to optimise the system, or 
they are imposed externally, or they could be uncertain and are described probabilistically.
Such parameter dependent problems have received widespread attention in recent
years, as evidenced by the survey \citem{BennWilcox-paramROM2015} and the work described in
e.g.\ \citem{BuiWilGhatt2008, DrohHass2012, Quarteroni2014, 
  MoRePaS2015, chinestaWillcox2017, GiraldiNouy2019}.
For some of the numerical procedures for producing reduced order models (ROMs),
e.g.\ \citem{Quarteroni2015, CohenEtal2015, HesthavenRozza2016, ChenSchwab2017}, it is of 
advantage if the parameter dependence appears in an affine or linear fashion.  {This can
be achieved by the methods of analysis described here.}  The parametrised
ROM is typically found through some kind of least squares principle, which is normally
equivalent to an orthogonal projection onto some approximating sub-manifold / subspace,
or by more general projection procedures.  Such parametrised problems can be analysed,
and aided in finding good approximations, by considering associated linear maps
\citem{hgmRO-1-2018, hgmRO-3-2019}. 
The parametric object in question  
is encoded in a linear map containing the same information in a different manner.
Such a linear map is then structurally a much more accessible object than a general parametric
object.  In particular, methods of analysis like the singular value decomposition (SVD) and 
spectral decomposition may be used to analyse them and thus understand approximations by 
low-dimensional parametric reduced order models (ROMs) in a different way.  The only
requirement imposed here is that for one parameter there is only one parametric object,
i.e.\ this is a mapping.  Otherwise no constraints are imposed on the set of parameters. 
Such formally parametric problems appear also
in the context of uncertainty quantification (UQ), and may be dealt with
in an essentially similar manner \citem{hgm-3-2018}.

Conditional expectation (CEX) on the other hand, which is the main ingredient in the Bayesian
treatment of inverse problems \citem{DashtiStuart2017}, also called Bayesian
updating (BU) \citem{BvrAkJsOpHgm11, hgmEzBvrAl:2016, hgmEzBvrAlOp2016, hgm17-ECM}
or data assimilation, is in its exact form \citem{Bobrowski2006/087}
also described as an orthogonal projection onto a subspace, or in the more general
form of a Fréchet mean onto a sub-manifold.  This is behind {basic forms of regression},
and also many ideas in filtering, such as the Kalman filter (KF), the Ensemble Kalman filter (EnKF)
\citem{Evensen}, and the Gauss-Markov Kalman filter (GMKF) \citem{bvrAlOpHgm12-a, opBvrAlHgm12}.
This approach, to use some kind of least squares principle, is also prevalent in practically
all forms of machine learning (ML) and neural network approximations, e.g.\
\citem{schmidh2014, Strang2019, SoizeGhanem2019, Schwab2019, RaissiKarniadakis2019, 
Fleuret2023, NelsenStuart2024}.
Its connection with probability and Bayesian updating has also long been recognised in the machine
learning (ML) community \citem{Murphy2012}.  
{Indeed, use of least-squares principles in approximations, regression, and order
reduction has a long tradition, going back at least to Gauss.  
Through the variational formulation of the conditional
expectation (CEX), this connection is made explicit.  This opens the possibility
of viewing or interpreting such least-squares procedures as an instance of a CEX, and
further of explicitly introducing additional, but often neglected, uncertainties into
the consideration when computing a parametric ROM.}
For more results concerning the connection
of ML or deep learning, and parametrised partial differential equations (PDEs) 
\citem{KutyniokEtal2019, FrescaEtal2020}.

The idea here is to combine these two approaches, and explicitly make the connection
between parametric ROM computation, Bayesian updating (BU), conditional expectation (CEX),
and machine learning (ML).  Connections between ROMs and data assimilation have
been explored also earlier \citem{BinevEtal2017}.
To this end, the developments in \citem{hgmRO-1-2018, hgmRO-3-2019}, {which
treat the encoding of parametric problems into linear maps},
will be briefly recalled in \fsec{S:lin-maps} for the convenience of the reader.
Likewise, in \fsec{S:CEX} a variational definition of the conditional
expectation, and its connection with projections, least squares, inverse problems,
and filtering is briefly sketched.
Finally, in \fsec{S:ROM-CEX} the two concepts are brought together to investigate
the connection between and interpretation of parametric ROMs and conditional
expectation.  This probabilistic interpretation then allows to introduce additional
uncertainties into consideration.  Conclusions are in \fsec{S:concl}.  Additionally,
appendices are attached which very briefly recall the various forms of the spectral 
decomposition of self-adjoint operators in \appendixname~\ref{S:spec-dec}, and
the singular value decomposition (SVD) in \appendixname~\ref{S:sing-val}.

\section{Analysis by Linear Maps}  \label{S:lin-maps}
Consider an operator equation, e.g.\ some physical {system} modelled and
depending on a {quantity} ${q}$:
\begin{equation}   \label{eq:evol-eq}
   \frac{\di}{\di t}u+ A({q};u) = f({q}) \qquad u\in \Vvk, \; 
   u(0)=u_0\in\Vvk;\;f \in \Vvk^*,
\end{equation}
where $\Vvk$ is the space of {states}, and $\Vvk^*$ is the dual
space of {actions / forcings}.
{The parameter} $p \in\C{P}$ may be
\[ p = q \quad | \quad p =(q, f) \quad | 
  \quad p=(q, f, u_0) \quad | \quad 
  p = (t, q,\ldots) \ldots , \]
  or  {time} $t$, {frequency} $\nu$, or
  a random {realisation} $\omega$, {coupling conditions} $\ldots$.
In total the solution $u(p)$ to \feq{eq:evol-eq} is dependent on the parameter
$p \in \C{P}$.

This is a typical example of a parameter dependent problem, which may be the
abstract formulation of a parameter dependent partial differential
equation (PDE).  It is what will be called the high fidelity model (HFM)
here, as an abstract mathematical model of a real-world solution with
a hopefully small \emph{modelling error}.

\begin{xmpn}[Aquifer]  \label{xmp:gr-water-1}
As a concrete example of a HFM of the abstract operator formulation in \feq{eq:evol-eq},
one may consider the governing equations according to Darcy's law for a ground-water flow 
problem in some spatial domain $\C{G} \subset \D{R}^d$ representing an aquifer,
with appropriate boundary and initial conditions:
\begin{equation}  \label{eq:evol-eq-PDE}
 \frac{\dd w}{\dd t} -\nabla \cdot \left( \vek{\kappa} \cdot \nabla w \right) = g,  
\end{equation}
where $w(x,t), x \in \C{G}$ is the hydraulic head, the symmetric second order positive definite 
spatial tensor field $\vek{\kappa}(x)$ is the hydraulic conductivity, $\nabla$ is the Nabla
operator, and $g(x,t)$ are sinks and sources.  Apart from the boundary and initial
conditions, one may regard as parameter $p$ the spatio-temporal field of sinks and sources 
$g(x,t)$, as well as the log-conductivity symmetric second order tensor field 
$\vek{q}(x) = \log \vek{\kappa}(x)$.  The $\log$ here is the inverse of the (matrix)
exponential, so that $\vek{\kappa}(x) = \exp \vek{q}(x)$.

Frequently, the hydraulic conductivity $\vek{\kappa}$ resp.\ the log-conductivity $\vek{q}$
is uncertain, and may be modelled as a random field $\vek{q}(x;\omega)$, where
$\omega_q \in \Omega_q$ describes the realisation in a probability space $\Omega_q$.  
Similarly, the sinks and sources may be uncertain, or they may partly describe extraction
of fluid by pumping, and thus a kind of control.  Therefore tey may be modelled as a random field
$g(x,t;\omega_g),\, \omega_g \in \Omega_g$, so that \feq{eq:evol-eq-PDE} is described
on the total probability space $\omega = (\omega_q, \omega_g) \in \Omega_q \times \Omega_g =: \Omega$
with total probability measure $\prob_{\Omega}$.

Possible parametric objects are for example the tensor field $\vek{q}(x;\omega_q)$ itself, the
right hand side field $g(x,t;\omega_g)$, the solution field, i.e.\ hydraulic head 
$w(x,t; \vek{q}, g) \equiv w(x,t; \omega)$, or even the differential operator 
$A(\omega): w \mapsto -\nabla \cdot \left( \vek{\kappa} \cdot \nabla w \right)$.
The parametric modelling of such second order positive definite spatial material tensor fields
is a subject in itself, for more information see e.g.\ \citem{ShaBrHgm24}.
\end{xmpn}

Often one is interested
in what will be termed a \emph{quantity of interest} (QoI), say
something like
\begin{equation}   \label{eq:evol-QoI}
   Y(u) = \D{A}(y(u(p),p)) ,
\end{equation}
where $\D{A}$ may be some kind of averaging process of a function
$y(u,p): \Uvk\times\C{P} \to \C{Y}$ over the set $\C{P}$,
which picks up values of the solution $u$ for specific values of the parameter $p$.
The situation where $\C{P}$ is a probability space and
$\D{A} = \Ex$ is the corresponding expectation operator is a
typical example.

\setcounter{xmpn}{0}
\begin{xmpn}[Aquifer --- cont.]  \label{xmp:gr-water-2}
In the aquifer \feX{xmp:gr-water-1}, imagine that there is a small domain $\C{G}_{e} \subset \C{G}$
where fluid is extracted.  This fluid has to flow in through the boundary $\dd \C{G}_e$.
Denoting at $x \in \dd \C{G}_e$ the outer normal by $n(x) \in \RR^d$, the inflow at that
point is $ -\nabla (\vek{\kappa}(x) w(x,t;\omega)) \cdot n(x)$, hence the total inflow is
$y(w(\omega),\omega) = \int_{\dd \C{G}_e} \, y(w(x,t;\omega))\, \di S(x)$, where $\di S(x)$ is the
surface differential.  Thus a possible QoI analogous to \feq{eq:evol-QoI} could be
the expected fluid extraction at time $t$:
\begin{equation}   \label{eq:evol-QoI-PDE}
   Y(w,t) = \Ex_{\Omega}(y(w(\cdot),\cdot)) = \int_{\Omega} \int_{\dd \C{G}_e} \, 
   y(w(x,t;\omega))\, \di S(x) \;\prob_{\Omega}(\di \omega),
\end{equation}
where $\Ex_{\Omega}$ is the expectation operator on $\Omega$.
\end{xmpn}

\subsection{The parametric map}   \label{SS:param-map}
As $A(p,\cdot)$, $f(p)$, $u(p)$, or $y(u(p),p)$ are all parameter dependent quantities,
we shall \emph{generically} refer to any of them as ${r}({p})$.
Assume that for each $p\in\C{P}$: $r(p)\in\Uvk$ is in a {linear} space,
and one wants to {represent} $p\mapsto r(p)$ computationally, i.e.\ find
{approximations} or reduced order models (ROMs) $r_a(p) \approx r(p)$.
To fix ideas, assume that $r(p)$ is explicitly given, or results from a general equation
--- like the evolution equation before --- written as
\begin{equation}   \label{eq:impl-fct-eq}
   F(r(p),p) = 0, \; \text{ {implicitly} defining }\; r(p);
\end{equation}
well-posed in the sense that for each $p\in\C{P}$ there is a
{unique} solution $r(p)\in\Uvk$ as implied by the {implicit function theorem}.

The parameter set $\C{P}$ is {not assumed to have any structure}, it will be replaced,
using {duality}, by a {vector subspace} {$\C{Q}$} of $\RR^{\C{P}}$ (the vector space
of real functions on $\C{P}$).
The main idea in what follows is then is to use this vector space structure 
\citem{hgmRO-1-2018, hgmRO-3-2019} of $\C{Q}$ and encode the parametric object in a 
linear map $R: \C{U} \to \C{Q} \subseteq \RR^{\C{P}}$ containing the same information.
For the sake of simplicity, it will be assumed that both  $\C{U}$  and $\C{Q}$
are Hilbert spaces.  The analysis and further developments then proceed from
the singular value decomposition (SVD, cf.\ \appendixname~\ref{S:sing-val}) of $R$
and the related spectral decomposition (cf.\ \appendixname~\ref{S:spec-dec}) of
$C_{\C{U}} = R^\dagger R : \C{U} \to \C{U}$, where $R^\dagger: \C{Q} \to \C{U}$
is the Hilbert space adjoint of $R$.  This leads to a representation of $r(p)$
on $\C{Q}$ with an expression linear resp.\ affine in the new parameters on $\C{Q}$.  
In addition, it will be shown that any other
\emph{factorisation} of $C_{\C{U}} = B^\dagger B$, where $B:\C{U} \to \C{H}$ is a
linear map into another Hilbert space $\C{H}$, leads to a linear representation on $\C{H}$,
so that every factorisation is connected to a linear representation, and vice versa.

\subsection{A Simple Example}   \label{SS:simpl-xmpl}
To make clear where we are going, take a look at a simple example:
assume that the set $\C{P}=\{p_1, p_2, p_3\}$ finite,
and $\vek{r}(p)=[r_1(p), r_2(p), r_3(p), r_4(p)]^\trpos\in\D{R}^4 = \Uvk$. 
 Arranging gives
\[ \vek{R} =
  \begin{bmatrix} 
     {r_1(p_1), r_2(p_1), r_3(p_1), r_4(p_1)} \\
     {r_1(p_2), r_2(p_2), r_3(p_2), r_4(p_2)} \\
     {r_1(p_3), r_2(p_3), r_3(p_3), r_4(p_3)} \\
  \end{bmatrix} ,
\]
a {matrix}, which corresponds to a {linear} map 
$\vek{R}: \Uvk=\D{R}^4 \to \D{R}^{\C{P}}$.

The {action} of $\vek{R}$ for any
$\vek{u} = [u_1, u_2, u_3, u_4]^\trpos \in \Uvk=\D{R}^4$ 
is given by
\[
   \vek{R} \vek{u} = \vek{r}(\cdot)^\trpos \vek{u} =
     [{\phi(p_1)}, {\phi(p_2)}, {\phi(p_3)}]^\trpos
     = \vek{\phi} \in \D{R}^{\C{P}} \cong \D{R}^3,
\]
where $\phi_k=\phi(p_k) = \sum_{n=1}^4 r_n(p_k) u_n$ --- a {weighted} average.
Obviously, knowing $\vek{R}$ and {knowing} $\vek{r}(p)$ is {equivalent}.
This is generally also true if $\C{P}$ is infinite, or $\Uvk$ is infinite dimensional.
In fact, in those situations the description by a linear map is actually
the theoretically more viable and more general concept, see e.g.\ \ \citem{hgm-3-2018}.
The advantage of dealing with $\vek{R}$ is that it is a linear map, where many
tools for analysis exist, whereas $p \mapsto \vek{r}(p)$ is just a not very well
characterised or specified function.

\subsection{Associated Linear Map}   \label{SS:ass-lin}
Thus, following \citem{hgmRO-1-2018, hgmRO-3-2019}, 
to {each} such function $r(p)$ with values in $\Vvk$, where for the 
sake of simplicity we assume that $\Vvk$ is a Hilbert space, one actually considers
\begin{equation} \label{eq:r-map-to-U}
r: \C{P} \to \Uvk := \cl \spn \im r = \cl \spn \{ r(p) \in \Vvk \mid p \in \C{P} \} \subseteq \Vvk,
\end{equation}
the map into the closed Hilbert subspace $\Uvk \subseteq \Vvk$ which is actually 
``hit'' by $r(p)$.  This will make the associated linear map
\begin{equation} \label{eq:r-map-to-U-2}
R: \Uvk \ni u  \mapsto  \bkt{r(p)}{u}_{\Uvk} \in \tilde{\C{R}} \subseteq \RR^{\C{P}}
\end{equation}
injective or one-to-one by design, where $\tilde{\C{R}} = \im R = R(\Uvk)$.  
More complicated situations, where $\Vvk$
is just a locally convex space,  may be dealt with
in an essentially similar manner \citem{hgm-3-2018} with some more effort.

One may use the injectivity of $R$ to make $\tilde{\C{R}}$ into a pre-Hilbert space
by transferring the inner product of $\Uvk$:
\begin{equation} \label{eq:inner-prod-RKHS}
  \forall \phi, \psi \in \tilde{\C{R}}:\; \bkt{\phi}{\psi}_{\C{R}}
:= \bkt{{R}^{-1}\phi}{{R}^{-1}\psi}_{\Uvk}.
\end{equation}
Denote the completion of $\tilde{\C{R}}$ with the resulting Hilbert norm as $\C{R}$,
then \feq{eq:inner-prod-RKHS} shows that ${R}^{-1}$ is unitary on this completion $\C{R}$,
which means that ${R}^{-1} = R^*$ and $R^* R = I_{\Uvk}$ and $R R^* = I_{\C{R}}$,
where $R^*$ is the adjoint of $R$,  $I_{\Uvk}$ is the identity on $\Uvk$,
and similarly for $I_{\C{R}}$.

\subsection{Reproducing Kernel Hilbert Space}   \label{SS:RKHS}
It is not difficult to check that $\C{R}$ is a \emph{reproducing kernel Hilbert space}
\citem{berlinet} (RKHS)  with {symmetric} positive definite \emph{reproducing kernel}
\begin{equation} \label{eq:kernel-RKHS}
  \vkappa(p_1,p_2) =  \bkt{r(p_1)}{r(p_2)}_{\Uvk}
 \in \D{R}^{\C{P}\times\C{P}}; \qquad \forall p \in \C{P}:\;
 \vkappa(p,\cdot) \in \C{R},
\end{equation}
 and the kernel function spans the whole space:
 $\overline{\spn} \{\vkappa(p,\cdot)\;|\;p \in \C{P}\} = \C{R}$; which shows that
 $\C{P}$ is represented in $\C{R}$ through $\C{P}\ni p \mapsto \vkappa(p, \cdot) \in \C{R}$.

The {reproducing} property,  which is the Riesz representation of the evaluation
function --- the \emph{Dirac}-$\updelta_p$ --- is expressed as
\begin{equation} \label{eq:kernel-repro}
  \forall \phi \in \C{R}:\; \bkt{\vkappa(p, \cdot)}{\phi(\cdot)}_{\C{R}} = \phi(p)= 
   \ip{\updelta_p}{\phi} = \updelta_p(\phi),
\end{equation}
where $\ip{\cdot}{\cdot}$ is the duality bracket between $\Uvk$ and its dual space.

In {other} settings (classification, machine learning, support vector machines),
when different subsets of $\C{P}$ have to be  {classified},
the space $\Uvk$ and the map  $r:\C{P} \rightarrow \Uvk$ is not given,
but can be freely {chosen} --- as {factorisation} of a kernel, cf.\ \feq{eq:kernel-RKHS}.
It is then called the {feature} map, and the whole procedure is called the {kernel trick}.

What is important is that up to now only the inner product on $\Uvk$ has been
used; it determines the one on the RKHS $\C{R}$ in \feq{eq:inner-prod-RKHS}.  
The ``correlation'' $R^* R = I_{\Uvk}$
is the identity --- see \fsec{SS:correlation} --- and can thus not distinguish
which part of $\C{P}$ is important.

But it is still possible to see that this construction allows a representation
which is linear resp.\ affine in the parameters.  To this end,
assume for the sake of simplicity that $\C{R}$ --- hence also $\Uvk$ --- 
is separable, and {choose} a {complete orthonormal system} ({CONS}) --- a Hilbert basis --- 
$\{\eta_m\}_m \subset \C{R}$ such that 
$\overline{\spn} \{\eta_1, \eta_2, \ldots\} = \C{R}$.

With $R^{-1}$ unitary, this gives a {CONS in $\Uvk$}:  $u_m = {R}^{-1} \eta_m$, and 
with that a {representation} of $r(p)$ on the RKHS $\C{R}$, one which
is \emph{linear} in these new parameters $\vek{\eta} = (\eta_1,\dots,\eta_m,\dots)$: 
\begin{equation} \label{eq:r-represent-RKHS}
r(p) = \sum_m \eta_m(p) u_m =: r(\vek{\eta}),
\end{equation}
as $R u_m = \bkt{r(p)}{u_m}_{\Uvk} = \eta_m(p)$ is the coefficient in the 
expansion of $r(p)$ in the CONS $\{u_m\}_m$.  Observe that in this situation
one can write the map $R$ and $R^*$ as
\begin{equation} \label{eq:R-represent-RKHS}
R = \sum_m \eta_m \otimes u_m, \; \text{ and } \; R^{-1}=R^* = \sum_m u_m \otimes \eta_m,
\end{equation}
a \emph{singular value decomposition} (SVD) of $R$ and $R^*$, 
see \appendixname~\ref{S:sing-val} --- all singular values and numbers are equal to unity.

The functions $\eta_m:\C{P}\to\D{R}$ can be considered as {``co-ordinates''} on 
$\C{P}$, or as  {new} {parameters} in which the representation $r(\vek{\eta})$ on the
RKHS $\C{R}$ is \emph{linear}.  Hence, one goal, that of a representation 
linear in the parameters, has been reached. 
Alternatively, one could have started with a {CONS} $\{u_m\}_m \subset \Uvk$, 
and then set $\eta_m = {R} u_m$ to achieve the same end.
But to find a \emph{good} reduced order model one needs a criterion on which
CONS is ``good'', so that an expansion like \feq{eq:r-represent-RKHS}
will give a good approximation with the first few terms.

\subsection{Correlation}   \label{SS:correlation}
To proceed towards the goal of finding a good CONS,
assume that $\C{Q} \subset \RR^{\C{P}}$ is another Hilbert space of 
real-valued functions on $\C{P}$ with an inner product $\bkt{\cdot}{\cdot}_{\C{Q}}$.
This inner product on $\C{Q}$ in some way ``measures'' what is {important} on 
$\C{P}$ resp.\ in $\D{R}^{\C{P}}$ --- e.g.\ if $(\C{P}, \mu)$ is a {measure} space, 
one may set $\C{Q} := \Lp_2(\C{P}, \mu)$.  Additionally,
assume that the linear map $R:\Uvk\to\C{Q}$ is densely defined and {closed};
although, again for the sake of simplicity, we shall mostly operate under the
assumption that $R$ is continuous / bounded, and hence can be defined on all of $\Uvk$.
 
The linear map is now $R: \Uvk \to \C{Q}$, but by slight abuse of notation we shall
keep the same denotation for it.   Its adjoint is now determined by the inner
product on $\C{Q}$, and to distinguish this from the situation in the
previous \fsec{SS:RKHS},  it will be denoted by $R^\dagger$;
and it will not be unitary any more.

To describe this situation with a new inner product in relation with the
linear map $R$, an additional linear map {$C_{\Uvk}$} --- the 
self-adjoint positive definite ``{correlation}'' --- 
can be defined by the {bilinear} form
\begin{equation} \label{eq:corr-U} 
\forall u,v \in \Uvk:\; \bkt{C_{\Uvk} u}{v}_{\Uvk} := \bkt{Ru}{Rv}_{\C{Q}}
=  \bkt{R^\dagger R u}{v}_{\Uvk}; \qquad R^\dagger \text{ w.r.t. } \C{Q},
\end{equation}
so that $C_{\Uvk} = R^\dagger R$.  In the case $\C{Q} = \Lp_2(\C{P}, \mu)$, one has
$C_{\Uvk} = \int_{\C{P}} r(p) \otimes r(p) \, \mu(\di p)$.

The singular value decomposition (SVD) 
--- see \appendixname~\ref{S:spec-dec} and \appendixname~\ref{S:sing-val} ---
of the adjoint map $R^\dagger : \C{Q} \to \Uvk$ is then the basis for a representation.
Assuming again for the sake of simplicity of exposition a 
discrete spectrum for the correlation $C_{\Uvk}$ in \feq{eq:corr-U}
with positive eigenvalues $\vsigma^2_j$ and eigenvectors $v_j \in \Uvk$,
i.e.\ a spectral decomposition $C_{\Uvk} = \sum_j \vsigma^2_j \, v_j \otimes v_j$
--- see \appendixname~\ref{S:spec-dec} ---
this gives for the SVD of $R^\dagger$ and the linear representation of $r(p)$
\begin{equation} \label{eq:lin-rep}
R^\dagger = \sum_j \vsigma_j \, v_j \otimes s_j, \quad
r(p) = \sum_j \vsigma_j \,  s_j(p) v_j .
\end{equation}
Here the $s_j\in \C{Q}$ are the normalised eigenfunctions belonging to the
eigenvalue $\vsigma^2_j$ of the associated ``companion'' correlation or kernel
$C_{\C{Q}} := R R^\dagger : \C{Q} \to \C{Q}$; recall that $C_{\C{Q}}$ has the 
same spectrum as $C_{\Uvk}$.  The action of $C_{\C{Q}}$ can still be described 
with the kernel function $\vkappa$ like in \feq{eq:kernel-repro}, but now 
with the $\bkt{\cdot}{\cdot}_{\C{R}}$-inner product there replaced 
by the $\bkt{\cdot}{\cdot}_{\C{Q}}$-inner product on $\C{Q}$; 
hence the kernel is \emph{not} reproducing any more. 
As may be gleaned from
the second member in \feq{eq:lin-rep}, the parametric quantity $r(p)$ is now again
represented linearly in the \emph{new parameters} $\vek{s} = (s_1, \dots, s_j, \dots)$,
i.e.\ $r(p) = r(\vek{s}(p))$.  But in contrast to \feq{eq:r-represent-RKHS}, now one 
has the singular values $\vsigma_j$ to indicate what terms in \feq{eq:lin-rep} are important.
The expansion of $r(p)$ in \feq{eq:lin-rep} is equivalent to the well known 
{\KL}-expansion (KLE) (see e.g.\ \citem{Matthies_encicl}), due to the {SVD} of $R^\dagger$.

Observe that $R$ (or  $R^\dagger$) leads to a {factorisation} of 
${C_{\Uvk} = R^\dagger R}\in\E{L}(\Uvk)$.  To {each} such factorisation belongs 
a representation / re-parametrisation of $r(p)$ and {vice versa} \citem{hgmRO-1-2018}.
Many other factorisations are
possible, and any two factorisations are unitarily equivalent in a certain
sense \citem{hgmRO-1-2018}.  They lead to other linear expansions similar to 
\feq{eq:lin-rep}, but we shall not dwell on this here and point the interested reader 
to \citem{hgmRO-1-2018, hgmRO-3-2019}, see also \appendixname~\ref{S:spec-dec}.

The development into more refined situations,
where e.g.\ the space $\Uvk = \Uvk_1 \otimes \Uvk_2$ is a tensor product,
or likewise $\C{Q} = \bigotimes_k \C{Q}_k$, and from where the SVD may be seen as
a low-rank tensor representation, will not be continued here for the sake of brevity.  
Again the interested reader is pointed towards
\citem{hgmRO-1-2018, hgm-3-2018, hgmRO-3-2019} and the literature cited therein.
Finally we point out that such separated or tensor representations are {intimately} connected 
with {deep} artificial neural networks, which can also be used for {functional approximation},
i.e.\ to approximate similar separated representations of $r(p)$ as \feq{eq:lin-rep},
see e.g.\ \citem{CohenSha2016, CohenShaEtal2018} and the references cited there.

\setcounter{xmpn}{0}
\begin{xmpn}[Aquifer --- cont.]  \label{xmp:gr-water-3}
Continuing the aquifer \feX{xmp:gr-water-1}, 
let us pick the solution, the hydraulic head
$w(x;\omega)$ in \feq{eq:evol-eq-PDE} as parametric object --- we have dropped the time-argument
$t$ to avoid unnecessary clutter of notation.  One may take as parameters
$\C{P} = \Omega$, and as Hilbert space $\C{Q} = \Lp_2(\Omega, \prob_{\Omega})$, the Hilbert
space of real random variables with finite variance, whereas for $\C{U}$ one may take
$\C{U} = \Lp_2(\C{G})$, the usual Hilbert space of real-valued square-integrable functions
on the domain $\C{G} \subset \RR^d$.  The mapping $R$ (cf.\ \feq{eq:r-map-to-U-2}) then reads
\begin{equation}  \label{eq:r-map-to-Q-PDE}
  R: \C{U} \ni v \mapsto \left(\psi_v: \omega \mapsto \psi_v(\omega) =
  \bkt{w(\cdot;\omega)}{v}_{\C{U}} =  (Rv)(\omega) \right) \in \C{Q},
\end{equation}
hence $(Rv)(\omega) = \psi_v(\omega) = \int_{\C{G}} w(y;\omega) v(y) \, \di y$.
The spatial correlation $C_{\C{U}}$ defined by the bilinear form \feq{eq:corr-U} satisfies
\begin{multline} \label{eq:corr-U-PDE} 
\forall u,v \in \Uvk:\; \bkt{C_{\Uvk} u}{v}_{\Uvk} := \bkt{Ru}{Rv}_{\C{Q}} =
\Ex_{\Omega}((Ru)(\cdot)(Rv)(\cdot)) \\
= \Ex_{\Omega}(\psi_u \psi_v )
= \iint_{\C{G} \times \C{G}} v(x) \vkappa_{\C{U}}(x,y) u(y) \, \di y \,\di x ,
\end{multline}
where the symmetric positive definite integral kernel is $\vkappa_{\C{U}}(x,y) = 
\Ex_{\Omega}(w(x;\cdot) w(y;\cdot))$.

The eigenfunctions $v_j\in \C{U} = \Lp_2(\C{G})$ may then be computed 
from the eigen-problem of the Fredholm integral
equation with kernel $\vkappa_{\C{U}}$ on $\C{G}$:
\begin{equation}   \label{eq:Fredholm-U}
   (C_{\Uvk} v_j)(x) = \int_{\C{G}} \vkappa_{\C{U}}(x,y) v_j(y)\, \di y =
   \vsigma_j^2 v_j(x) .
\end{equation}
\end{xmpn}

\subsection{Reduced Order Models}   \label{SS:ROMs}
The example problem \feq{eq:evol-eq} can typically not be solved directly
and has to be discretised in some way, a process in which the space $\Vvk$
is replaced by a finite dimensional subspace $\Vvk_n \subseteq \Vvk$,
and the equation in \feq{eq:evol-eq} is replaced by its discretised version,
what will be termed the \emph{full order model} (FOM) here:
\begin{equation}   \label{eq:evol-eq-FOM}
   \frac{\di}{\di t}\vek{u}+ \vek{A}(p;\vek{u}) = \vek{f}(p), \qquad \vek{u} \in \Vvk_n.
\end{equation}
The QoI \feq{eq:evol-QoI} can then be evaluated on the FOM solution
$\vek{u}(p)$:
\begin{equation}   \label{eq:evol-QoI-FOM}
   Y(\vek{u}) = \D{A}(y(\vek{u}(p),p)) ,
\end{equation}
which hopefully is close to $Y(u) \approx Y(\vek{u})$.

Thus in general one looks at quantities $\vek{r}: \C{P} \to \Uvk_n \subset \Uvk$,
where $\dim \Uvk_n = n$ is typically a large number.
The preceding developments leading to \feq{eq:lin-rep} can now be repeated
for the FOM quantities with a map $\vek{R}: \Uvk_n \to \C{Q}$, yielding
the analogue of \feq{eq:lin-rep}:
\begin{equation} \label{eq:lin-rep-FOM}
\vek{R}^\trpos = \sum_j \vsigma_j \, \vek{v}_j \otimes s_j, \quad
\vek{r}(p) = \vek{r}(\vek{s}(p)) = \sum_{j=1}^n \vsigma_j \,  s_j(p) \vek{v}_j,
\quad \vek{s} = [s_1,\dots,s_n]^\trpos;
\end{equation}
where the $\vek{v}_j \in \Uvk_n, j=1,\dots,n,$ are now eigenvectors of 
$\vek{C}_{\Uvk,n} = \vek{R}^\trpos \vek{R}$, and the functions $s_j(p) \in \C{Q}$
are the eigenvectors of $\vek{C}_{\C{Q},n} = \vek{R} \vek{R}^\trpos$.

From \feq{eq:lin-rep-FOM} one can very easily build a reduced order model (ROM).
Assume, as it is customary (cf.\ \appendixname~\ref{S:sing-val}), that the
singular values $\vsigma_j$ are ordered by decreasing magnitude.  By
truncating the sum in \feq{eq:lin-rep-FOM} at some number $M \ll n$,
one has a ROM
\begin{equation} \label{eq:lin-rep-ROM}
\vek{R}^\trpos_M = \sum_{j=1}^M \vsigma_j \, \vek{v}_j \otimes s_j, \quad
\vek{r}_M(p)  = \vek{r}_M(\vek{s}(p)) = \sum_{j=1}^M \vsigma_j \,  s_j(p) \vek{v}_j;
\end{equation}
where the neglected terms produce an error which may be measured by
$\sum_{j>M} \vsigma_j$.
Obviously, the range 
\begin{equation} \label{eq:UM-ROM-im}
\im \vek{R}^\trpos_M = \vek{R}^\trpos_M(\C{Q}) = \spn \{\vek{v}_1,\dots,\vek{v}_M\}
=: \Uvk_M \subset \Uvk_n \subseteq \Uvk 
\end{equation}
is a $M$-dimensional subspace, hence $\vek{R}_M: \Uvk_M \to \C{Q}$.
One may now compare $\vek{C}_{\Uvk,M} := \vek{R}^\trpos_M \vek{R}_M$ to 
$\vek{C}_{\Uvk,n}$, and this to $C_{\Uvk}$, to gauge the quality of the ROM.

Similarly, in case some vectors $\{\vek{u}_1,\dots, \vek{u}_N\} =: \Uvk_a \subset \Uvk$
are picked by some procedure, defining a finite-dimensional subspace $\Uvk_a$ on which to
approximate $\vek{r}(p)$ by a separated representation ROM 
\begin{equation} \label{eq:UM-ROM}
\vek{r}_a(p)  = \vek{r}_a(\vmu(p)) = \sum_{j=1}^N \mu_j(p) \vek{u}_j    
\approx \vek{r}_n(\vmu(p)) \approx r(p), 
\end{equation}
where we do not go into details on how the vectors $\vek{u}_j$ are picked nor how
the coefficient functions $\vmu = [\mu_1,\dots,\mu_N]^\trpos$ are computed (e.g.\ consult
\citem{BennWilcox-paramROM2015}, or
\citem{BuiWilGhatt2008, DrohHass2012, Quarteroni2014, 
  MoRePaS2015, Quarteroni2015, CohenEtal2015, HesthavenRozza2016, 
  chinestaWillcox2017, ChenSchwab2017, GiraldiNouy2019} for some possibilities), 
one can again use the above procedure and define a map $\vek{R}_a: \Uvk_a \to \C{Q}$ by
\begin{equation} \label{eq:R-ROM}
\Uvk_a \ni \vek{w} \mapsto \vek{R}_a \vek{w} := 
   \bkt{\vr_a(p)}{\vek{w}}_{\Uvk} = \vr_a(p)^\trpos \vek{w} \in \C{Q}.
\end{equation}
Again, one may form $\vek{C}_{\Uvk,a} := \vek{R}_a^\trpos \vek{R}_a$ 
and compare it to $\vek{C}_{\Uvk,n}$ and $C_{\Uvk}$ to gauge the 
quality of the ROM.  More generally, any approximation $\vek{r}_b(p) \approx \vek{r}(p)
\approx r(p)$, separated like \feqs{eq:UM-ROM}{eq:R-ROM} or not,
defines its linear map $R_b$, and again one may form 
$\vek{C}_{\Uvk,b} := \vek{R}_b^\trpos \vek{R}_b$
to study the approximation properties on the easier subject of linear maps.

The forms \feqss{eq:lin-rep-FOM}{eq:lin-rep-ROM}{eq:R-ROM} allow a slight reformulation:
one regards the quantity $\vr(p) \approx \vr_M(\vek{s}(p)) \approx \vr_a(\vmu(p))$
not as a function of $p\in \C{P}$, but rather as functions of $\vek{s} \in \RR^N$ or
$\vmu \in \RR^N$.  Taking the last case, one then looks at a mapping
\begin{equation} \label{eq:mu-ROM}
\Uvk_{a} \ni \vek{w} \mapsto \vek{R}_{\mu} \vek{w} := 
   \bkt{\vr(\vmu)}{\vek{w}}_{\Uvk} = \vr(\vmu)^\trpos \vek{w} \in \Mlg,
\end{equation} 
where $\Mlg$ is an appropriate $N$-dimensional 
Hilbert space of real-valued functions in $\vmu \in \RR^N$;
and again, one may form $\vek{C}_{\Uvk,\mu} := \vek{R}_\mu^\trpos \vek{R}_\mu$ for
further computations, and especially 
$\vek{C}_{\Mlg,\mu} := \vek{R}_\mu \vek{R}^\trpos_\mu: \Mlg \to \Mlg$.

It is possible to go even a step further, and, by choosing a basis in in the $N$-dimensional 
vector space $\Uvk_{a}$, replace it by $\RR^N$, and the parametric object $\vr_{\mu}(\vmu)$ 
hence by $\vr_N(\vmu) = [r^1_N(\vmu),\dots,r^N_N(\vmu)]
= \sum_{j=1}^N r^j_N(\vmu) \ve_j$. 

Then \feq{eq:mu-ROM} becomes
\begin{equation} \label{eq:N-ROM}
\RR^N \ni \vek{w} \mapsto \vek{R}_{N} \vek{w} := 
   \bkt{\vr_{N}(\vmu)}{\vek{w}}_{\RR^N} = \vr_N(\mu)^\trpos \vek{w} 
   = \sum_{j=1}^N r^j_N(\vmu) w^j \in \Mlg.
\end{equation} 
 Now, by also choosing 
a basis in $\im \vek{R}_{N} = \vek{R}_{N}(\RR^N)$ --- where, as $R$ as injective, so is
$\vek{R}_{N}$, and hence $\dim (\im \vek{R}_{N}) = N$ and thus $\vek{R}_{N}(\RR^N) \cong \RR^N$ ---  
one obtains that the operators $\vek{R}_{N}$, $\vek{R}_{N}^\trpos$ 
are represented as $N \times N$ matrices.
As before, one forms the correlations --- also represented as $N \times N$ matrices ---
$\vek{C}_{N,\mu} := \vek{R}_N^\trpos \vek{R}_N$ and
$\vek{C}_{\Mlg,\mu} := \vek{R}_N \vek{R}^\trpos_N: \Mlg \to \Mlg$.

\setcounter{xmpn}{0}
\begin{xmpn}[Aquifer --- cont.]  \label{xmp:gr-water-4}
Continuing further the aquifer \feX{xmp:gr-water-1}, the HFM in \feq{eq:evol-eq-PDE} will usually
be discretised, say by the finite element method, or, as it makes no difference for the
considerations here, more generally by a Galerkin method to obtain a FOM like in \feq{eq:evol-eq-FOM}.
Of interest here is that actually the forms \feqss{eq:lin-rep-FOM}{eq:lin-rep-ROM}{eq:R-ROM}, 
as well as \feqs{eq:mu-ROM}{eq:N-ROM}, are all similar as regards our overview.  
For the computation one chooses a basis as in \feq{eq:N-ROM}, and
the operators $\vek{R}_{N}$, $\vek{R}_{N}^\trpos$, and $\vek{C}_{N,\mu}$, $\vek{C}_{\Mlg,\mu}$,  
are actually represented by $N \times N$ matrices, cf.\ \fsec{SS:ROM-RF}.

This means that the  eigen-problem of the Fredholm integral
equation in \feq{eq:Fredholm-U} is a linear algebra problem of computing
the spectral decomposition of a  matrix
\begin{equation}   \label{eq:matrix-U}
   \vek{C}_{N,\mu} \vv_j = \vsigma^2_j \vv_j .
\end{equation}
Similar remarks apply to the eigen-problem of $\vek{C}_{\Mlg,\mu}$, which has the
same eigenvalues as \feq{eq:matrix-U}.  From both of these eigen-problems one can piece
together the SVD of $\vek{R}_{N}$ and $\vek{R}_{N}^\trpos$, which of course can also
be computed directly via standard linear algebra procedures.
\end{xmpn}

One should point out once more that the establishment of a HFM \feq{eq:evol-eq} involves
a modelling error, an uncertainty; the discretisation to a FOM \feq{eq:evol-eq-FOM}
invokes the discretisation error, another uncertainty.  Transitions to a ROM
of whatever kind triggers an approximation error, an additional uncertainty.
To deal with uncertainties is one of the uses of probability theory, and this
will lead to the subject of conditional expectation.

\section{Conditional Expectation}   \label{S:CEX}
The notion of conditional expectation (CEX) is normally based on the concept of
conditional probability, which is also the background of Bayes's theorem.
These concepts are basic to the probabilistic solution of \emph{inverse} problems,
and in our brief summary we follow \citem{hgmEzBvrAl:2016, hgmEzBvrAlOp2016, hgm17-ECM}. 
Such inverse problems may arise when one wants to determine e.g.\ the value of the
parameter $p\in\C{P}$ in \feq{eq:evol-eq-FOM} by observing the function $y(u,p)$
which appears in the evaluation of the QoI \feq{eq:evol-QoI}.  Knowing $\check{y} = y(u(p_o),p_o)$,
one would like to determine $p_o \in \C{P}$, i.e.\ to \emph{invert} the map
$p \mapsto y(u(p),p)$.  Unfortunately, these observation maps are typically not invertible,
the values $\check{y}$ do usually not contain enough information to determine $p_o$,
and the inverse problem is therefore typically \emph{ill-posed}, and thus can not be
approached directly in a numerical fashion.

It will be advantageous to look at conditional expectation from a variational
perspective, an approach which is not so common in the literature.  To see how
this comes about, and how it is connected to the common treatment of conditional
probabilities, a short excursion to the theorem of Bayes and Laplace and the
problems connected with it is taken.

Typically, a parametric ROM tries to match the performance of a FOM as accurately as
possible, and the measure of this is usually some kind of least-squares expression to
be minimised.  This expression can be labelled as a \emph{loss-function}. 
  As an example, one may regard the solution of a FOM at some
parameter values as the new information, and additionally one may regard the mathematical
form of the parametric ROM as given information --- or alternatively as an approximation for
the subspace characterising the FOM.  Machine learning (ML) normally proceeds very
similarly,  cf.\ e.g.\
\citem{schmidh2014, Strang2019, SoizeGhanem2019, Schwab2019, RaissiKarniadakis2019, 
Fleuret2023, NelsenStuart2024}.

It turns out that the conventional conditional expectation is equivalent to an orthogonal
projection onto a subspace, which in some way contains the new information on which
one is conditioning.  This means that according to Pythagoras
it minimises the square of the distance resp.\ difference 
between a random variable and its CEX --- i.e.\ a least-squares solution.
It is now not too difficult to see that this may be seen in the context of ROM computation
as just another interpretation of normal computational procedures.  The change of language and point
of view lead to a formulation in terms of random variables and probability measures.  In the
ML community this view is actually quite common.

But now that one has taken this probabilistic point of view, one may go beyond
of just a new interpretation of what one is doing anyway.  One may in fact widen the 
approach as a Bayesian update via conditional expectation (CEX) and
include the uncertainties of errors which arise from various approximation steps --- like going 
from reality to HFM to FOM to ROM --- in a probabilistic way.  These errors and their 
uncertainties were already addressed in  \fsec{SS:ROMs} and
will be considered in more detail in \fsec{SS:ROM-UQ}.  Including these uncertainties
in the Bayesian update via CEX when determining a parametric ROM allows one to minimise
the \emph{total loss} as measured by by a distance squared in a larger space accomodating these
additional uncertainties, and not just the ones resulting from the transition FOM to ROM, 
as is customary.  One may now go further in a Bayesian fashion and consider the risk of
subsequent decisions taken on the basis of using some ROM results instead of te --- typically at
that point still unknown --- reality.  This may well lead to a
modification of the loss-function away from least-squares to better reflect the risks in 
the decision.  This then is not a conditional expectation anymore, but may actually
minimise the risks which
come from using the results of a ROM instead of reality for decisions taken.

\subsection{The Theorem of Bayes and Laplace}  \label{SS:bayes}
By embedding the inverse problems in a probabilistic setting 
(e.g.\ \citem{jaynes03} and references therein), it usually
becomes  \emph{well-posed} \citem{hgmEzBvrAlOp2016, Latz2023}.  
This comes about by considering
$p \in \C{P}$ as a random variable (RV) --- the so-called \emph{prior} --- 
and not wanting to recover a particular $p_o$, but rather a 
\emph{probability distribution} on $\C{P}$ --- the so-called \emph{posterior} --- 
which signifies how probable it is that a particular $p \in \C{P}$ is 
the sought element $p_o \in \C{P}$.  The change in the probabilistic 
description to the posterior model comes from additional information on 
the system through measurement or observation --- like $\check{y}$.

Formally, assume that the uncertain parameters are given by
\begin{equation}  \label{eq:RVp}
{p}: \Omega_p \to  \C{P}  \text{  as a RV on a probability space   }
  (\Omega_p, \F{A}_p, \D{P}_p) ,
\end{equation}
where the set of elementary events is $\Omega_p$, $\F{A}_p$ a $\sigma$-algebra of
measurable events, and $\D{P}_p$ a probability measure.  Additionally, also
the situation / action / loading / experiment may be uncertain, and we model
this by allowing also $f\in\Vvk^*$ in \feq{eq:evol-eq} to be a random variable
on some probability space $(\Omega_f, \F{A}_f, \D{P}_f)$. 
The \emph{expectation} or mean of a RV
corresponding to $\D{P}_p$ will be denoted by $\EXP{\cdot}_p$,
e.g.\ $\bar{p}:=\EXP{{p}}_p := \int_{\Omega_p} {p}(\omega_p) \, \D{P}_p(\di \omega_p)$,
and the zero-mean part is denoted by $\tilde{p} = {p} - \bar{p}$.  The covariance
of ${p}$ and another RV ${q}$ is written as 
${C}_{pq} := \EXP{\tilde{p}\otimes\tilde{q}}_p$, and for short ${C}_{p}$ if ${p}={q}$.

Since $p$ and $f$ in \feq{eq:evol-eq} are RVs, so is $u(p)$, and in the end also the
prediction of the  ``true'' observation $y(u(p),p)$ in the QoI \feq{eq:evol-QoI}.
Additionally, it is commonly assumed that there is an observational uncertainty,
given by a $\C{Y}$-valued RV $\vepsilon(\omega_e)$ defined on a probability
space $(\Omega_e, \F{A}_e, \D{P}_e)$, so that the observed $\check{y}$ is a sample
from $z := y(u(p),p) + \vepsilon$.  
For simplicity, the three sources of uncertainty are often considered as \emph{independent},
so that on the total probability space $\Omega = \Omega_p \times \Omega_f \times \Omega_e$
the probability measure is a product measure $\D{P} = \D{P}_p \otimes \D{P}_f \otimes \D{P}_e$.
 
Bayes's theorem is commonly accepted as a consistent way to incorporate
new knowledge into a probabilistic description \citem{jaynes03}.
The elementary textbook statement of Bayes's theorem by Laplace is about
conditional probabilities
\begin{equation}  \label{eq:iII}
 \D{P}(\C{I}_p|\Mlg_z) = \frac{\D{P}(\Mlg_z|\C{I}_p)}{\D{P}(\Mlg_z)}\D{P}(\C{I}_p),
 \quad \text{ if }\;\D{P}(\Mlg_z)>0,
\end{equation}
where $\C{I}_p\subset\C{P}$ is some subset of possible ${p}$'s on which we would like
to gain some information, and $\check{y} \in \Mlg_z\subset\C{Y}$ is the information
provided by the measurement.  The term $\D{P}(\C{I}_p)$ is the so-called
\emph{prior}, it is what we know before the observation $\Mlg_z$.
The quantity $\D{P}(\Mlg_z|\C{I}_p)$ is the so-called \emph{likelihood},
the conditional probability of $\Mlg_z$ assuming that $\C{I}_p$ is given.
The term $\D{P}(\Mlg_z)$ is the so called \emph{evidence}, the probability
of observing $\Mlg_z$ in the first place, which sometimes can be expanded
with the \emph{law of total probability}, allowing one to choose between
different models of explanation.  This term is necessary to make the
right hand side of \feq{eq:iII} into a real probability --- summing
to unity --- and hence the term $\D{P}(\C{I}_p|\Mlg_z)$, the so-called \emph{posterior}
which reflects our knowledge on $\C{I}_p$ \emph{after} observing $\check{y} \in \Mlg_z$.

The statement in \feq{eq:iII} runs into problems if the set of observations
$\Mlg_z$ has vanishing probability measure, $\D{P}(\Mlg_z)=0$, as is the case 
when we observe a \emph{continuous} random variable $z$ and $\Mlg_z$ is a 
one-point set.  The statement \feq{eq:iII} then has the indeterminate
term $0/0$, and some form of limiting procedure is needed.  As a sign that
this is not so simple --- there are different and inequivalent forms of doing
it --- one may just point to the so-called \emph{Borel-Kolmogorov} paradox, 
see e.g.\ \citem{Pollard2001, jaynes03}.

The theorem in \feq{eq:iII} may in the $\D{P}(\Mlg_z)=0$ case be
formulated in \emph{densities} under certain restrictive conditions, or more 
precisely in probability density functions (pdfs).  This is the case 
when $z$ and ${p}$ have a \emph{joint} pdf $\pi_{z,p}(z,{p})$.
Then Bayes's theorem \feq{eq:iII} may be formulated as 
\begin{equation}  \label{eq:iIIa}
 \pi_{p|z}({p}|z) = \frac{\pi_{z,p}(z,{p})}{Z_s(z)},
\end{equation}
where $\pi_{p|z}({p}|z)$ is the \emph{conditional} pdf, and $Z_s$
(from German \emph{Zustandssumme} --- ``sum over all states'')
is a normalising term such that the conditional pdf $\pi_{p|z}(\cdot|z)$
integrates to unity;
\[ Z_s(z) = \int_{\Omega}  \pi_{z,p}(z,{p}) \, \D{P}(\di {p}) . \]
The joint pdf may be split into the \emph{likelihood density} 
$\pi_{z|p}(z|{p})$ and the \emph{prior} pdf $\pi_p({p})$
\[ \pi_{z,p}(z,{p}) =  \pi_{z|p}(z|{p}) \pi_p({p}) . \]
so that \feq{eq:iIIa} has its familiar form
\begin{equation}  \label{eq:iIIa1}
 \pi_{p|z}({p}|z) = \frac{\pi_{z|p}(z|{p})}{Z_s(z)}  \pi_p({p}) .
\end{equation}
These terms correspond directly with those in \feq{eq:iII} and carry the same names.

In case one can establish a conditional measure $\D{P}(\cdot|\Mlg_z)$
or even a conditional pdf $\pi_{p|z}(\cdot|z)$,
the \emph{conditional expectation} (CEX) $\EXP{\cdot|z}$ may be
defined as an integral over that
conditional measure resp.\ the conditional pdf.  Thus, originally, 
the conditional measure or pdf implies the conditional expectation \citem{jaynes03}.
But the question remains on how to use conditioning in the cases
where Bayes’s original formula is not applicable.  This was solved by Kolmogorov
\citem{Bobrowski2006/087} and is the subject of the next 
section \fsec{SS:Var-CEX}, and will at the same time introduce 
the variational view on conditioning which will be used later in \fsec{S:ROM-CEX}.

\subsection{Variational Statement of Conditional Expectation}  \label{SS:Var-CEX}
The problem sketched in the previous \fsec{SS:bayes} by conditioning on events 
of zero probability
\citem{jaynes03, Bobrowski2006/087}
was solved by turning things around, and taking as primary notion not
conditional probability as in \feq{eq:iII}, or even conditional pdfs as in \feq{eq:iIIa},
but to first define \emph{conditional expectation} (CEX), conditioned on a 
\emph{sub-$\sigma$-algebra} $\F{B} \subset \F{A}$ of a probability space $(\Omega,\F{A},\D{P})$.  
Kolmogorov's definition of a CEX used the Radon-Nikodým derivative, but for our
purposes it is most naturally approached in the $\Lp_2$-setting as a least-squares
procedure.
For two RVs $\vphi, \psi$, define their
inner product by $\bkt{\vphi}{\psi}_2 := \EXP{\vphi \psi}$, and set
$\Svk := \Lp_2(\Omega,\F{A},\D{P}) = \{ \vphi | \bkt{\vphi}{\vphi}_2 < \infty \}$
to be the Hilbert space of (equivalence classes of) RVs with finite $\Lp_2$-norm
$\nd{\vphi}_2 := \sqrt{\bkt{\vphi}{\vphi}_2}$.  The Hilbert space formed with a
sub-$\sigma$-algebra $\F{B}$ is a closed subspace, denoted by
$\Svk_\infty := \Lp_2(\Omega,\F{B}) \subset \Lp_2(\Omega,\F{A})$. 
Denote the one-dimensional subspace of constants generated by the 
sub-$\sigma$-algebra $\F{B}_0 := \{ \Omega, \emptyset \}$ as 
$\Svk_0 := \spn \{ \bbbone_{\Omega}\}$, where $\bbbone_{\Omega}$ is the
function or RV with constant value equal to unity on the set $\Omega$.

To see the motivation behind the next definition and what we are aiming
at in the case of ROMs, observe first that the normal expectation of a RV
$x \in \Lp_2(\Omega,\F{A}) = \Svk$, given the inner 
product and norm on $\Svk$, can be defined in a variational way as
\begin{equation}  \label{eq:EX-def}
  \EXP{x} = \arg \min_{\chi \in \Svk_0}  \Psi_{x}(\chi) = P_{0}x \in \Svk_0,
\end{equation}
where $\Psi_{x}(\chi) := \nd{x - \chi}_2^2 = \EXP{(x - \chi)^2}$ 
is the $x$-based \emph{variance}, and $P_0$ is the orthogonal projection onto $\Svk_0$.
This function is also known the Bayesian \emph{loss-function}, which the 
(conditional) expectation minimises.  In this context the requirement
or condition that $\chi \in \Svk_0$, i.e.\ that it is a constant function 
$\chi = \alpha \bbbone_{\Omega}$ for some $\alpha \in \RR$, may be seen as the 
``prior'' information.

On such a Hilbert subspace given by a sub-$\sigma$-algebra $\F{B}$, 
one may now give a variational characterisation \citem{Bobrowski2006/087} for the conditional
expectation $\Ex(x | \F{B})$ of a RV $x \in \Lp_2(\Omega,\F{A})$ w.r.t.\ that
sub-$\sigma$-algebra $\F{B}$:
\begin{equation}  \label{eq:CEX-def}
  \Ex(x | \F{B}) := \Ex(x | \Svk_\infty) := \arg \min_{\chi \in \Svk_\infty} \Psi_{x}(\chi)
  = P_{\F{B}} x \in \Svk_\infty,
\end{equation}
where we recall that $\Svk_\infty = \Lp_2(\Omega,\F{B})$.
One may immediately observe that this definition, which requires the minimisation of a 
continuous quadratic --- a strictly convex functional --- over a closed subspace, yields
a unique RV $\Ex(x | \F{B}) \in \Lp_2(\Omega,\F{B})$, and that 
$\Ex(x | \F{B}) = P_{\F{B}} x$ is the \emph{orthogonal projection} ---
denoted by $P_{\F{B}}$ --- of the RV $x$ onto the closed subspace $\Lp_2(\Omega,\F{B})$.

This implies a version of the theorem of Pythagoras:
\begin{equation}  \label{eq:CEX-Pyth}
  \nd{x}_2^2 = \nd{x - \chi}_2^2 + \nd{\chi}_2^2,
\end{equation}
as well as the \emph{Galerkin} orthogonality condition
\begin{equation}  \label{eq:CEX-orth}
  \forall \chi \in \Lp_2(\Omega,\F{B}): \bkt{x - \Ex(x | \F{B})}{\chi}_2 = 0,
\end{equation}
or equivalently, for all $\chi \in \Lp_2(\Omega,\F{B})$ one has 
$\EXP{x \chi} = \EXP{\Ex(x | \F{B}) \chi}$.  Note that this last statement can 
then be extended to all $\Lp_1$-functions by re-interpreting it in the 
$\Lp_1 - \Lp_\infty$ duality.

 For an observation such as the one generated by
the RV $z = y(x) + \vepsilon$, one would normally take as sub-$\sigma$-algebra 
$\F{B}$ the one generated by these RVs.  Thus in this case
\[
  \Lp_2(\Omega,\F{B}) = \{ \vphi \in \Lp_2(\Omega,\F{A}) \mid \vphi = \chi(z) \;
     \text{ is a function of } \; z=y(u,p) + \vepsilon \},
\]
i.e.\ all RVs which are functions of $z=y(x) + \vepsilon$ and are in $\Lp_2$.
Thus for such a special sub-$\sigma$-algebra $\F{B}$ one has
\begin{equation}  \label{eq:CEX-fct}
  \Ex(x | \F{B}) = \phi_{x}(z)
\end{equation}
for some measurable function $\phi_x(\cdot)$.
This RV $\phi_{x}(z)$ gives the CEX as a function
of all possible observations $z = y(x) + \vepsilon$.  After
an observation $\check{y}$, one may then compute the \emph{constant}
\begin{equation}  \label{eq:CEX-post}
\Ex(x | \check{y}) := \phi_{x}(\check{y}),
\end{equation}
which we like to call the \emph{conditioned} expectation or post-CEX.

So, for each RV $x \in \Svk = \Lp_2(\Omega,\F{A})$, one has an assignment
like \feq{eq:CEX-post}, which is linear in $x$ as it results from the orthogonal 
projection $P_{\F{B}} x$ in \feq{eq:EX-def}, and it easy to see that
$\EXP{\bbbone_{\Omega} | \F{B}} = P_{\F{B}} \bbbone_{\Omega} = \bbbone_{\Omega}$,
so that $\Ex(\bbbone_{\Omega} | \check{y}) = \phi_{\bbbone_{\Omega}}(\check{y}) = 1$.
This means that $\Ex(x | \check{y})$ in \feq{eq:CEX-post} satisfies all
requirements one has for an expectation operator, which allows one
to define a conditional probability, as one sets for all measurable $\C{E} \subseteq \Omega$
\begin{equation}  \label{eq:CEX-prob-meas}
   \D{P}(\C{E} | \check{y}) := \Ex(\bbbone_{\C{E}} | \check{y}),
       \quad \C{E} \in \F{A}.
\end{equation}

The variational definition may be shown \citem{Bobrowski2006/087} to yield
the classical definitions \feq{eq:iII} of Bayes and Laplace when those are applicable.
It also allows two generalisations, which we briefly mention.  For one,
if the CEX is used in some decision process, one might incorporate the resulting
gains or losses measuring the \emph{risk} resulting from this decision under uncertainty,
into the Bayesian loss-function --- like in \feqs{eq:EX-def}{eq:CEX-def} --- 
and hence minimise some notion of \emph{risk}.
Secondly, the definition \feq{eq:CEX-def} can be extended to more general
metric spaces than Hilbert spaces: For a metric space $\C{Z}$ with metric
$\delta$ and a $\C{Z}$-valued RV $\xi$ one may broaden the Bayesian loss-function to 
something like $\Psi_{\delta, \xi}(\chi) := \EXP{\delta(\xi, \chi)^2}$ to
minimise over all $\C{Z}$-valued RVs to arrive at the so-called
\emph{Fréchet mean} and Fréchet conditional expectation.

\subsection{Approximations of Conditional Expectation and Filtering}  \label{SS:filter-CEX}
It is worthwhile to recall which information went into establishing the post-CEX
\feq{eq:CEX-post}.  It on one hand obviously the observation $\check{y}$ of the RV $z$.
But it is also on the other hand the subspace $\Svk_\infty = \Lp_2(\Omega,\F{B})$
which is generated by \emph{all} reasonable functions of the observed quantity $z$.
So this is what might be called ``prior'' knowledge: for one the \emph{expectation operator}
$\Ex$ which defines the probabilistic description of the RV $x$, but also the 
subspace $\Svk_\infty$, here generated by all functions of the observed quantity $z$.
This defines the conditional expectation (CEX) \feq{eq:CEX-def}.  The second ingredient
is the observation $\check{y}$ of the RV $z$, from which one may then compute the 
\emph{conditioned expectation}  \feq{eq:CEX-post}.

In any attempt to actually compute the CEX from \feq{eq:CEX-def} one immediately
notices that in any realistic situation the subspace $\Svk_\infty$
is in some sense too ``big'' and difficult to handle numerically.  This suggest the
necessity of approximating \feq{eq:CEX-def} by choosing a $m$-dimensional subspace  
$\Svk_m \subseteq \Svk_\infty$ of $\Svk_\infty$ for the minimisation.  
After all, this is also what is behind the original 
Bubnov-Galerkin idea of computing approximations to elliptic PDEs.  One can interpret
this also as some kind of additional prior information,
in that we ``know'' that $\phi_{x} \in \Svk_m$, i.e.\ that $\phi_{x}$ is to be found
in the subspace $\Svk_m$.

Behind many filtering concepts is the idea of not changing the probability measure
$\D{P} \mapsto \D{P}(\C{E} | \check{y})$ resp. \ $\EXP{\cdot} \mapsto \EXP{\cdot | \check{y}}$,
but to change the RV $x_f$ --- the subscript $f$ now means ``forecast'' --- to another
one $x_a$ --- here the subscript $a$ now means ``assimilated'', as it has \emph{assimilated}
the new information --- i.e.\ $x_f \mapsto x_a$, such that
\begin{equation}  \label{eq:fore-ass}
   \Ex(\vphi(x_f) | \check{y}) = \Ex(\vphi(x_a))
\end{equation}
for any reasonable measurable function $\vphi$.  The \feq{eq:fore-ass} means that
any descriptor --- given by the function $\vphi$ --- of the RV $x_f$ according to 
the new probability assignment $\Ex(\cdot | \check{y})$ is equal to the same 
descriptor of the RV $x_a$ according to the original probability assignment $\Ex(\cdot)$.
These two ways of looking at BU may be compared to the Heisenberg and Schrödinger
views on quantum mechanics evolution; in the Heisenberg picture it is the state
which is evolving, i.e.\ the probability measure resp.\ expectation operator
is changing while the observables (another word for RVs) stay as they are, 
whereas in the Schrödinger picture it is the other way around.  Hence the
filtering idea corresponds to the Schrödinger picture.

To use the CEX in Bayesian updating (BU), as the first step one might want
to construct a filter which will at least get the CEX for the identity right,
i.e.\ $\Ex(x_f | \check{y})$.  So one starts from the orthogonal decomposition
underlying Pythagoras's theorem in \feq{eq:CEX-Pyth}:
\begin{equation}  \label{eq:CEX-ortho-dec}
  x_f = (x_f - P_{\F{B}} x_f) + P_{\F{B}} x_f = (I - P_{\F{B}}) x_f + P_{\F{B}} x_f .
\end{equation}
As the observation $\check{y}$ of $z$ contains information about the last term 
$P_{\F{B}} x_f \in \Svk_\infty$ in \feq{eq:CEX-ortho-dec}, one may proceed
\citem{hgmEzBvrAl:2016, hgmEzBvrAlOp2016, hgmLitZ16-x, hgm17-ECM}, by modifying 
that last term in \feq{eq:CEX-ortho-dec} from $P_{\F{B}} x_f = \Ex( x_f | \F{B})$
to $\Ex( x_f | \check{y})$, and thus set for the filter equation:
\begin{equation}  \label{eq:CEX-filter}
  x_a^{\text{CEX}} := (x_f - \Ex( x_f | \F{B})) + \Ex( x_f | \check{y}) = 
     x_f + (\Ex( x_f | \check{y}) - \Ex( x_f | \F{B})) = x_f + x_i ,
\end{equation}
where $x_i = \Ex( x_f | \check{y}) - \Ex( x_f | \F{B})$ is called the \emph{innovation}.  
As \feq{eq:CEX-orth} implies that $\Ex( x_i ) = 0$, it is clear that 
$\Ex(x_a^{\text{CEX}}) = \Ex( x_f | \check{y})$, so that the new assimilated RV
$x_a^{\text{CEX}}$ has as expected value the correct conditional expectation of $x_f$.

But the exact minimisation over $\Svk_\infty$ to obtain $\Ex( x_f | \F{B})$ is still
typically not possible, so further approximations are necessary even for \feq{eq:CEX-filter}.
Picking up the idea to approximate $\Svk_\infty$, 
one of the simplest possible approximations in \feq{eq:CEX-def} is to take instead
$\Svk_1 = \{ \vphi \in \Svk \mid \vphi(z) = Lz + b \bbbone_{\Omega} \}$, 
i.e.\ the space of affine functions of $z$.  The solution is not difficult to compute
and is essentially given by the well known Gauss-Markov theorem:
the solution
\begin{equation}  \label{eq:CEX-1}
  \EXP{x_f | \Svk_1 } := \arg \min_{\chi \in \Svk_1} \Psi_{x_f}(\chi)  \in \Svk_1,
\end{equation}
results from
minimising $\Psi_{x_f}(\cdot)$ over $\Svk_1$, and is $\phi_{x,1}(z) = K z + a$ with
$K := C_{x_f z} C_{z}^{-1}$ and $a := \bar{x} - K(\bar{z})$, where 
where $C_{x_f z}$ is the covariance of $x_f$ and $z$, and $C_z$ is the auto-covariance of $z$.
The linear operator $K$ is also called the \emph{Kalman gain}, and has the 
familiar form known from least-squares projections.  

Merging the CEX-filter \feq{eq:CEX-filter} together with the linear approximation
in \feq{eq:CEX-1} for a simple BU, one arrives at the Gauss-Markov-Kalman-filter (GMKF)
\citem{bvrAlOpHgm12-a, opBvrAlHgm12, hgmEzBvrAl:2016, hgmEzBvrAlOp2016, hgmLitZ16-x, hgm17-ECM}:
\begin{equation}  \label{eq:GMK-filter}
  x_a^{\text{GMKF}} := x_f + K(\check{y} - z) .
\end{equation}
As the GMKF operates on the usually infinite dimensional space $\Svk = \Lp_2(\Omega,\F{A})$,
it is still not readily computable and needs further discretisation.  One should recall
that using  \feq{eq:GMK-filter} just for the means,  
$\bar{x}_a^{\text{KF}} := \bar{x}_f + K(\check{y} - \bar{z})$,
plus an update for the covariance $C_{x x}$ gives the original Kalman filter (KF).
One can show \citem{bvrAlOpHgm12-a, opBvrAlHgm12} that the Kalman covariance update
is automatically satisfied by \feq{eq:GMK-filter}, as it operates on RVs.
Discretisations of \feq{eq:GMK-filter} have been done via the Monte Carlo (MC) method,
yielding \citem{Evensen} the Ensemble Kalman filter, one of the simplest \emph{particle filters},
where \feq{eq:GMK-filter} is sampled, and the covariances are computed from the samples,
and the MC samples are also called ``particles''.  Another discretisation is possible
by truncating polynomial chaos expansions of the RVs involved, and applying \feq{eq:GMK-filter}
on the truncated polynomial chaos expansions, or, equivalently, on the coefficients of the
polynomial chaos expansion \citem{bvrAlOpHgm12-a, opBvrAlHgm12}, giving the polynomial chaos
Kalman filter (PCKF).  The covariances are in this instance computed from the truncated
polynomial chaos expansions.

If in \feq{eq:CEX-1} larger subspaces $\Svk_m \supseteq \Svk_1$ are
used, one arrives at non-linear filters \citem{hgmEzBvrAl:2016, hgmEzBvrAlOp2016, hgm17-ECM}.
It is also possible to design procedures \citem{hgmLitZ16-x}, which, by computing the
CEX of functions $\vphi(x_f)$ of $x_f$, will create a filter which recovers the RV
$x_a$ in the Schrödinger picture with arbitrary accuracy, i.e.\ such that \feq{eq:fore-ass}
is satisfied by as many functions $\vphi$ as possible.  The hereby necessary
CEX computations can be performed with high accuracy \citem{JVhgm2018}
on the basis of the Galerkin orthogonality statement \feq{eq:CEX-orth}.

Looking towards the building of ROMs, some of the
methods --- which originally come from geo-statistics --- are built on the
idea of predicting a Gaussian process from samples, where the covariance of
the process is known.  This is essentially using the
expression \feq{eq:GMK-filter}, assuming that the covariances are known,
leading to \emph{Kriging} or \emph{Gaussian process emulation} 
\citem{Kennedy01, OHaganWest2010}.  To judge these procedures, one should point out
that for the situation where all RVs involved in \feq{eq:GMK-filter} are Gaussians,
the BU in \feq{eq:GMK-filter} is \emph{exact}; and as a Gaussian 
is fully described by mean and covariance, the original Kalman filter is also \emph{exact}.
Thus, using these Kriging procedures also in the non-Gaussian case can then be seen as 
using a linear approximation to BU to compute a ROM.  Similar methods as Kriging
essentially also use \feq{eq:GMK-filter}, but employ kernels which do not have
to come from covariances of random fields, but rather from RKHS (cf.\ \fsec{SS:RKHS})
or radial basis functions.

\section{Parametric ROMs and conditional expectation}    \label{S:ROM-CEX}
As was already alluded to at the end of the preceding \fsec{SS:filter-CEX}, there
are considerable similarities between Bayesian updating (BU) via linear filtering
and building proxy models using kernels, where both types of methods use variants
of \feq{eq:GMK-filter}.  In this case it is fairly obvious to draw conclusions on the
connection between reduced order models (ROMs) and conditional expectation (CEX).

Since the time of Gauss, and possibly even before that, the least-squares approach
was seen as a good method to do regression, leading to the well known Gauss-Markov theorem. 
And regression is one of the more obvious connections with
conditional expectation.  While there are methods to address inverse ill-posed problems
in a deterministic manner, in the ML and deep artificial neural network
community \citem{Murphy2012, schmidh2014, Strang2019, Fleuret2023} the ``learning task'' 
has always been seen in a probabilistic light, e.g.\ \citem{Neal1995}.  Therefore, here these
learning methods will not be addressed further, as they are already formulated in a Bayesian
probabilistic setting.  Here the focus is rather on the traditionally deterministic
formulations --- one could say deterministic learning methods --- to find ROMs, as the
combination with the probabilistic Bayesian ideas sheds new light on these methods, and
also allows one to consider additional uncertainties in a probabilistic sense.

\phantom{page} 

\subsection{Uncertainties in ROMs}  \label{SS:ROM-UQ}
One of the advantages of a probabilistic view, which were already mentioned, is that
other uncertainties, which appear in the whole process, can be incorporated in
the picture.  In \fsec{SS:ROMs} some of the different uncertainties were listed.
These are:
\begin{itemize}
\item The uncertainties inherent in the system $A$ in \feq{eq:evol-eq}, 
      described by a probability space $(\Omega_{p_r},\sigalg_{p_r},\prob_{p_r})$.
      Here the ${p_r}$ are the part of the parameters which describes unresolved uncertainties
      in the system.  The part of the parameters which we want to use and explicitly control 
      somehow will be denoted by $\vmu$, like in \feq{eq:UM-ROM}.  The
      uncertainties in the external action $f$ in \feq{eq:evol-eq} will be
      described by a probability space $(\Omega_f,\sigalg_f,\prob_f)$.
\item The \emph{modelling error} in going from the ``real system'' to its mathematical
      model (HFM) \feq{eq:evol-eq} resp.\ \feq{eq:impl-fct-eq}.  One way to capture this
      in the simplest case is by adding an \emph{additive} error either to the evolution equation
      \feq{eq:evol-eq} or to its solution $u(p)$.  In the case of the implicit description
      \feq{eq:impl-fct-eq}, one could again in the simplest case add an error to the equation
      \feq{eq:impl-fct-eq}, or to $r(p)$.  In any case, this uncertainty can be described
      with a probability space $(\Omega_{\text{HFM}},\sigalg_{\text{HFM}},\prob_{\text{HFM}})$.
\item The \emph{discretisation error} in going from the HFM \feq{eq:evol-eq} resp.\ 
      \feq{eq:impl-fct-eq} to the discretised full-order model (FOM) \feq{eq:evol-eq-FOM}.
      Such errors are usually described in the numerics community by the worst case situation,
      but for the purpose of this exposition, one might consider this again, like in the
      previous modelling error situation, to be described by an additive random variable
      on the probability space $(\Omega_{\text{FOM}},\sigalg_{\text{FOM}},\prob_{\text{FOM}})$.
\item The \emph{numerical error} which is due to the numerical solution --- down to some tolerance
      on finite precision hardware --- of the FOM system like \feq{eq:evol-eq-FOM}, or the same
      equation interpreted with the
      ROM approximations \feqs{eq:lin-rep-ROM}{eq:UM-ROM}.  Again, this error is usually only
      described in the numerics community by the worst case situation, but one might again consider
      it to be described by an additive random variable
      on the probability space $(\Omega_{\text{NUM}},\sigalg_{\text{NUM}},\prob_{\text{NUM}})$.
      This point is particularly relevant in the light of recent developments where one considers
      computations with less precision than the up to now usual 64 bit floating point format,
      in order to speed up the computation.  Some ideas even point in the direction of using
      \emph{variable} precision in different parts of the algorithm.
\item The \emph{reduction error} in going from the FOM \feq{eq:evol-eq-FOM} to the
      reduced order model (ROM) \feqs{eq:lin-rep-ROM}{eq:UM-ROM}.  
      This is the only error which is usually considered in a ROM computation.
      The uncertainty inherent in this error can again be
      described by an additive  random variable on the probability space 
      $(\Omega_{\text{ROM}},\sigalg_{\text{ROM}},\prob_{\text{ROM}})$.  In case one interprets
      the ROM computation as a CEX, then at least two components of the reduction error
      can be named more clearly.  One is the error which represents the ``observation error''
      $\epsilon$ of the observed variable $z = y + \epsilon$ of the observable $y$ in 
      \feq{eq:CEX-fct}, modelled on a probability space 
      $(\Omega_\epsilon,\sigalg_\epsilon,\prob_\epsilon)$.  The other is the error arising
      from approximating the CEX, e.g.\ as in \fsec{SS:filter-CEX}, which introduces an error
      modelled  on the probability space 
      $(\Omega_{\text{CEX}},\sigalg_{\text{CEX}},\prob_{\text{CEX}})$.
      Thus the probability space $\Omega_{\text{ROM}}$ can be envisaged as 
      $\Omega_{\text{ROM}} = \Omega_\epsilon \times \Omega_{\text{CEX}} \times \Omega_r$,
      where $\Omega_r$ represents some possible left-over residual error in $\Omega_{\text{ROM}}$.
\end{itemize}

When describing the solution to \feq{eq:evol-eq-FOM} with an additive error, this means that 
e.g.\ instead of dealing with $\vek{u}$, one deals with $\vek{u} + \vek{\eta}$, 
or, more generally instead of having the
parametric quantity $\vek{r}(\vmu)$ in the form of \feqs{eq:lin-rep-ROM}{eq:UM-ROM},
one deals with $\vek{r} + \vek{\eta}$.


For the sake of simplicity, let us combine all the uncertainties inherent in the HFM
to a new random variable $\vek{\eta}_M(\omega_M)$ defined on the probability space
$\Omega_M := \Omega_{p_r} \times \Omega_{f} \times \Omega_{\text{HFM}}$ with
a probability measure $\prob_M$ on a $\sigma$-algebra $\sigalg_M$ of subsets of $\Omega_M$,
with $\omega_M = (\omega_{p_r}, \omega_f, \omega_{\text{HFM}})$.  
Similarly, combine \emph{all} the additional numerical uncertainties
from the transitions HFM  $\to$ FOM, FOM  $\to$ ROM, ROM $\to$ numerical approximation,
i.e.\ all uncertainties
inherent in the numerical ROM solution of \feq{eq:evol-eq-FOM}, or in the parametric quantity
as in \feqs{eq:lin-rep-ROM}{eq:UM-ROM}, with a new random variable
$\vek{\eta}_N(\omega_N)$, defined on the probability space $\Omega_N := 
\Omega_{\text{FOM}} \times \Omega_{\text{ROM}} \times \Omega_{\text{NUM}}$, equipped with
a probability measure $\prob_N$ on a $\sigma$-algebra $\sigalg_N$ of subsets of $\Omega_N$,
with $\omega_N = \omega_{\text{FOM}} \times \omega_{\text{ROM}} \times \omega_{\text{NUM}}$.

It has to be pointed out that, although these different errors were described individually, they
do not have to be stochastically independent, e.g.\ the probability measure describing \emph{all}
the uncertainties from numerical processes in the solution of the ROM, $\prob_N$ on $\Omega_N := 
\Omega_{\text{FOM}} \times \Omega_{\text{ROM}} \times \Omega_{\text{NUM}}$, 
does not have to be a product measure $\prob_N = \prob_{\text{FOM}} \otimes \prob_{\text{ROM}}
 \otimes \prob_{\text{NUM}}$.  But it is not unreasonable to assume that the modelling error
described on $\Omega_M$, and the numerical error described on $\Omega_N$ are independent.
In that case one has that the total probability measure $\prob = \prob_{M} \otimes \prob_{N}$
is a product measure on the total probability product space 
$(\Omega = \Omega_M \times \Omega_N,\sigalg,\prob)$, 
and expectations w.r.t.\ $\prob$ can be computed via Fubini's theorem:
$\Ex = \Ex_N \circ \Ex_M = \Ex_N(\Ex_M(\cdot))$.

Finally, this means that instead of the ROM solution $\vek{u}$ to \feq{eq:evol-eq-FOM},
one deals with the RV $\vek{u}_{{MN}} = \vek{u} + \vek{\eta}_M + \vek{\eta}_N$.
Or, more generally, instead of having the
parametric quantity $\vek{r}(\vmu)$ in the form of \feqs{eq:lin-rep-ROM}{eq:UM-ROM},
one has to deal with $\vek{r}_{{MN}}(\vmu,\omega_M,\omega_N) = 
\vek{r}(\vmu) + \vek{\eta}_M(\omega_M) + \vek{\eta}_N(\omega_N)$.

As an example of this implicit acknowledgement of a probabilistic interpretation of
such errors, one may look at global climate simulations as they are reported e.g.\ in the 
Intergovernmental Panel on Climate Change Working Group's 
(IPCC WG I) contribution to the 6th Assessment Report \citem{IPCC2021-WG1-full}.
As best estimates for the climate evolution predicted from different modelling groups,
which differ from each other due to different modelling errors $\vek{\eta}_M(\omega_M)$
and different numerical errors $\vek{\eta}_N(\omega_N)$, the average
of these simulations is taken there as the ``best'' predictor.  This arose initially by just 
noticing that this average was a better predictor in the past, and apparently implicitly 
assumes that the different models form some kind of Monte Carlo sample of the 
variables $\vek{\eta}_M$, $\vek{\eta}_N$.

\setcounter{xmpn}{0}
\begin{xmpn}[Aquifer --- cont.]  \label{xmp:gr-water-5}
Further in the aquifer \feX{xmp:gr-water-1}, one may now try and obtain a probabilistic
model for the errors involved in representing reality by the HFM with the help
of the governing equation for Darcy flow.  This step will usually require specialised
domain knowledge, as can for example be found in the IPCC panel just mentioned as concerns
climate change.  In areas where controlled experiments are possible, one may use the results
of these in comparison with computations to estimate such errors and uncertainties.
One difficulty here is obviously that one can not solve the HFM directly, and one has to
turn to numerical approximations.  This immediately brings up the second group of errors.

The second group of errors is purely numerical, and involves everything in discretising
the HFM to the FOM, and further reducing to a parametric ROM.  As already mentioned,
these errors are usually treated in numerical analysis on a worst-case basis with bounds.
It is only recently that sometimes a probabilistic point of view was adopted.  For the
aquifer this would mean addressing the finite element discretisation error and further the
parametric ROM approximation error in a probabilistic setting.
\end{xmpn}

\subsection{Generalising conditional expectation}  \label{SS:re-inter-CEX}
To address the main topic in a unified way, recall that the conditional 
expectation (CEX) of a random variable (RV) $x$ is a minimisations of a ``loss functional''
$\Psi_x$ over some manifold $\Svk_g$.  In fact, in \feqs{eq:EX-def}{eq:CEX-def} one could 
see that the expectation and CEX are defined in a very similar way, the only difference
being the manifold $\Svk_g$.  One could actually argue that all expectations are 
\emph{conditional} expectations.  The general expression looks like
\begin{equation}  \label{eq:gen-CEX}
    \Ex_g(x | z) = \arg \min_{\chi \in \Svk_g(z)} \Psi_x(\chi)   ,
\end{equation}
where the manifold $\Svk_g(z) \subset \Svk$ 
is somehow determined by what is going to be the observation $z$.
In most cases, and the only ones we consider here, the functional is the squared distance,
$\Psi_x(\chi) = \nd{x - \chi}^2$, in which case $\Ex_g(x | z)$ is a projection onto 
$\Svk_g(z)$, i.e.\ a \emph{least-squares} solution, cf.\ \fsec{S:CEX}.
In the case of the normal expectation \feq{eq:EX-def}, one has $\Svk_g = \Svk_0 = 
\spn\{ \bbbone_{\Omega}\}$, i.e.\ the space of constants, 
whereas in the case of the CEX one has $\Svk_g(z) = \Svk_\infty = 
\{ \phi \in \Lp_2(\Omega) \mid \phi = \chi(z), \chi \text{ measurable }\}$,
i.e.\ practically all possible functions of $z$.

As was already alluded to in \fsec{SS:filter-CEX}, it is advantageous for our purpose to
discern in \feq{eq:gen-CEX} what is the \emph{prior}, i.e.\ what do we know before making
any observation, measurement, or evaluation of the RV $x$, and what is the \emph{posterior}.
In \feq{eq:gen-CEX}, what is known a priori is the RV $x$ --- or some function of $x$ to compute
some descriptor of $x$, but in our case the main interest is in $x$ itself, i.e.\ when the function
is the identity --- and what we are going to observe, i.e.\ the RV $z$.  Additionally what we
know a priori is the structure of the manifold $\Svk_g$, as the examples \feqs{eq:EX-def}{eq:CEX-def}
show.  And one knows a priori the type of Bayesian loss function $\Psi_x$, as well as the
probability space $\Svk = \Lp_2(\Omega,\sigalg,\prob)$ on which everything is 
described or modelled, cf.\ \fsec{SS:ROM-UQ}.

Together with the actual observation $\check{y}$, this then gives the new state or \emph{conditioned}
expectation functional $\Ex_g(\cdot | \check{y})$, as described previously in \fsec{S:CEX};
this updated expectation functional describes the \emph{posterior} probability description of
any RV, especially also $x$.  Our main interest is in the updated mean $\Ex_g(x | \check{y})\in\C{X}$,
where $\C{X}$ is the Hilbert space in which $x$ and $\chi$ live.
Consideration of all the uncertainties addressed above as new random variables 
in the end results in a larger probability space with more parameters, 
and thus forces the numerical computations to occur in higher dimensions.
In the expectation operator inherent in the functional in \feq{eq:gen-CEX}, one then uses
this generalised expectation operator, i.e.\
$\Psi_{x}(\chi) = \nd{x - \chi}^2_{\Svk} = \Ex_g(\nd{x - \chi}_{\C{X}}^2)$, where
$\nd{\cdot}_{\C{X}}$ is the Hilbert norm of $\C{X}$, and the least-squares solution
is now performed by averaging additionally over the new random variables.

According to the results \feq{eq:UM-ROM} in \fsec{SS:ROMs},
the ROMs can all be put in the form
$ \vr_a(\vmu) = \sum_{j=1}^N \mu_j \vek{u}_j  = \vr_a(\vmu)  \approx r(\vmu) $, 
where $ \vmu = (\mu_1,\dots,\mu_N), \mu_j \in \C{Q} $ may be seen as real-valued coordinates 
on the parameter set $\C{P}$, and the ROM $ \vr_a(\vmu)$ may be regarded as a function of
the new parameters $\vmu \in \RR^N$.  This view will be the basis of consideration from
here on.  

Thus one considers $x = \vr(\vmu)$ in \feq{eq:gen-CEX}, with $\C{X} = \Uvk$, 
and observations $y$ resp.\ $z = y + \vepsilon$ which
are some functions $z = \phi(\vr)$ of $\vr(\vmu)$; often linear functionals, 
e.g.\ samples $\vr^k \in \Uvk$ as measurement for $\vr(\vmu_k)$. 
The observation is then simply $z^k = \vr^k + \vek{\eta}_M(\omega_k)  + \vek{\eta}_N(\omega_k) 
\in \Uvk$.
 
The functional used in \feq{eq:gen-CEX} thus becomes
\begin{equation}   \label{eq:psi-r}
\Psi_{\vr}(z) = \nd{\vr(\vmu) - \chi(z)}^2_{\Svk} = 
   \Ex_g(\nd{\vr(\vmu,\cdot) - \chi(z,\cdot)}^2_{\Uvk}),
\end{equation}
where we take the expectation operator $\Ex_g$ which covers all the uncertainties as described
in \fsec{SS:ROM-UQ} in $\vr(\vmu,\omega)$ and $\chi(z(\vr,\omega),\omega)$.  
The minimisation in \feq{eq:gen-CEX} is carried out over
$\Svk_g$, an a priori specified set of functions of 
$z(\vr,\omega) = y(\vr,\omega) + \vepsilon(\omega)$,
which is peculiar to the specific ROM computation which is desired, so that the
ROM $\vr_a(\vmu)$, according to \feqs{eq:gen-CEX}{eq:psi-r}, may be seen as the
CEX of $\vr(\vmu)$:
\begin{equation}   \label{eq:ROM-CEX-g}
   \vr_a(\vmu) = \Ex_g(\vr(\vmu) | z) .
\end{equation}
This will be made a bit more concrete in the following sections,
by specifying the observations $z$ and the manifold $\Svk_g$.  One advantage 
which arises from this ``conditional expectation view'', and which concerns
any ROM calculation which falls into this mould, is the possibility to include
of all the uncertainties alluded to in \fsec{SS:ROM-UQ} in the computation of 
the desired ROM $\vr_a(\vmu)$.

The view taken here, which coincides with current statistical or econometric thinking,
is that  practically any least-squares procedure
can be seen as a conditional expectation or approximation thereof.  This is also
particularly true of (deep) artificial neural networks (e.g.\ \citem{Strang2019, Fleuret2023}).
This is even true for more general regressions, based on more general \emph{loss-functions}.
The least squares method is also used in the updating of a \emph{proper orthogonal decomposition}
(POD) basis \citem{chinestaWillcox2017}, which therefore can be seen also as a form of
conditional expectation.  The same can be said of the alternating minimisation
procedures employed in the computation of the \emph{proper generalised decomposition} (PGD),
an extension of the proper orthogonal decomposition, cf.\ e.g.\ \citem{falco-nouy-2011}.

\subsection{Parametric ROM as a Random Field}  \label{SS:ROM-RF}
According to the results in \feqs{eq:UM-ROM}{eq:mu-ROM}, one may computationally
regard the parametric model $\vr(\vmu)$ as a defined on $\vmu \in \RR^N$.
As $\vr(\vmu)$ lives in a $N$-dimensional vector space $\Uvk_a$, we replace
it with a $\RR^N$ by choosing a $\Uvk$-orthonormal basis in it, cf.\ \feq{eq:N-ROM}.  
Now the parametric model may be regarded as a map $\vr_N:\, \RR^N \ni \vmu 
\mapsto \vr(\vmu) \in \RR^N \cong \Uvk_a$, given by $\vr_N(\vmu) = \sum_{j=1}^N
r^j_N(\vmu) \ve_j$.  The $N$-dimensional function space $\Mlg$, the image of
$\vR_N: \RR^N \ni \vw \mapsto \vr_N(\vmu)^\trpos \vw \in \Mlg$, is spanned by the
$N$ functions $r^j_N(\vmu)$ $(j=1,\dots,N)$, linear in $\vmu$.

For the sake of simplicity, assume that the inner product on the space $\Mlg$
of $\vmu \in \RR^N$ functions is given by a measure $\rho$: $\bkt{\vphi}{\psi}_{\Mlg} := 
\int_{\RR^N} \vphi(\vmu)\psi(\vmu)\, \rho(\di \vmu)$. 
Then the correlation on $\Uvk \cong \RR^N$
according to \feqs{eq:corr-U}{eq:N-ROM}  in \fsec{SS:ROMs} is given by the matrix
\begin{equation}   \label{eq:corr-U-RN}
   \vC_{N,\mu} = \vR_N^\trpos \vR_N = \int_{\RR^N} \vK_N(\vmu,\vmu) \, \rho(\di \vmu) ,
\end{equation}
where the ``local correlation matrix'' $\vK_N$ is given by
$\vK_N(\vmu_1,\vmu_2) := \vr_N(\vmu_1) \otimes \vr_N(\vmu_2) = \vr_N(\vmu_1) \vr_N(\vmu_2)^\trpos$.
The corresponding correlation on $\Mlg$ is
\begin{equation}   \label{eq:corr-M-RN}
   \vC_{\Mlg,\mu} = \vR_N \vR_N^\trpos: \Mlg \ni \psi 
   \mapsto \int_{\RR^N} \vkappa_{N}(\cdot,\vmu) \psi(\vmu) \, \rho(\di \vmu) \in \Mlg ,
\end{equation}
where the kernel $\vkappa_{\vmu}$ (cf.\ \feq{eq:kernel-RKHS}) is
\begin{equation} \label{eq:kernel-on-mu}
  \vkappa_{N}(\vmu_1,\vmu_2) =  \bkt{\vr_N(\vmu_1)}{\vr_N(\vmu_2)}_{\Uvk} = \vr_N(\vmu_1)^\trpos
  \vr_N(\vmu_1)  \in \RR^{N \times N}; 
\end{equation}
which is the trace of the local correlation matrix used in \feq{eq:corr-U-RN}:
$\vkappa_{N}(\vmu_1,\vmu_2) = \tr \vK_N(\vmu_1,\vmu_2)$.

One way to look at the whole situation, especially in light of all the uncertainties
discussed in \fsec{SS:ROM-UQ} and \feq{eq:psi-r}, is, for the purpose of establishing a ROM, 
to regard $\vr_N(\vmu)$ as a \emph{random field} on $\RR^N$ with values in $\Uvk$.
Now all the procedures to estimate random fields --- usually based on conditional expectation ---
are at our disposal.  

Here only a very simple version will be sketched, namely the simplest case of the aforementioned
technique of Kriging or Gaussian process emulation (GPE) (e.g.\ \citem{Kennedy01, OHaganWest2010}).
To start, assume that $r_s(\vmu)$ is a zero-mean \emph{scalar random field} on $\RR^N$.  
The covariance kernel or function, according to \feq{eq:kernel-on-mu}, is also assumed to be known:
$\vkappa_s(\vmu_1,\vmu_2) = \Ex(r_s(\vmu_1) r_s(\vmu_2))$.  In practice, often the covariance function
is \emph{not known}, and assumed in a computationally convenient form.
Now assume in addition, that $r_s(\vmu)$ is a Gaussian process, 
hence completely characterised by (the vanishing mean and) the covariance function.  

Assume that the process has been observed at various locations $\vmu_1,\dots,\vmu_m$, with values
grouped as $\vz = [r_s(\vmu_1),\dots,r_s(\vmu_m)]^\trpos \in \RR^m$.  
To obtain a new generic location 
$\vmu \in \RR^N$, we assume the estimate $\chi_s(\vmu)$ to be a regression on $\vz$,
i.e.\ to be in the linear manifold $\Svk_{\text{GPE}} = \{ \chi_s(\vmu) \mid \chi_s(\vmu) =
\vw(\vmu)^\trpos \vz, \; \vw(\mu) \in \RR^m \} $. Hence, following \feq{eq:gen-CEX}, we set
\begin{equation}  \label{eq:GEP-CEX}
   \hat{r}_s(\vmu) = \Ex_{\text{GPE}}(r_s(\vmu) | \vz) = \arg \min_{\Svk_{\text{GPE}}(\vz)}
     \Psi_{r_s}(\chi_s) \in \Svk_{\text{GPE}}(\vz)  ,
\end{equation}
where $\Psi_{r_s}(\chi_s) = \Ex((r_s(\vmu) - \chi_s(\vmu))^2) = \var(r_s(\vmu) - \chi_s(\vmu))$.
It is a fairly standard computation to find the minimising $\chi_s(\vmu)$ resp.\ the minimising
weight vector $\vw(\vmu)$.  
Setting $\vK_{\text{GPE}} := (\vkappa_s(\vmu_i,\vmu_j))_{i,j=1}^m \in \RR^{m \times n}$ and
$\vg(\vmu) = [\vkappa_s(\vmu_1,\vmu),\dots,\vkappa_s(\vmu_m,\vmu)]^\trpos$, one obtains
\begin{equation}   \label{eq:Kriging-w}
  \vw(\vmu) = \vK_{\text{GPE}}^{-1} \vg(\vmu)\quad \text{ and } \quad \hat{r}_s(\vmu) =
      \vg(\vmu)^\trpos \vK_{\text{GPE}}^{-1} \vz.
\end{equation}

It is now not difficult to see that the procedure in \feq{eq:Kriging-w} can be performed
for every component of $\vr_N(\vmu) = [r_N^1(\vmu),\dots,r^N_N(\vmu)]^\trpos$, even
taking into account non-zero correlations between the components of $\vr_N(\vmu)$,
contained in the correlation matrix $\vK_N(\vmu_1,\vmu_2) := \Ex(\vr_N(\vmu_1) \otimes \vr_N(\vmu_2))
= \Ex(\vr_N(\vmu_1) \vr_N(\vmu_2)^\trpos)$.

Now the observations at the locations $\vmu_1,\dots,\vmu_m$ are
$\vZ = [\vz_1^\trpos,\dots,\vz_m]^\trpos]^\trpos \in \RR^{(Nm)}$ with $\vz_j = \vr_N(\vmu_j) \in \RR^N$, 
and the linear manifold is given by $\Svk_{N,\text{GPE}}(\vZ) = 
\{ \vek{\chi}_N(\vmu) \mid \vek{\chi}_N(\vmu) =
\vW(\vmu)^\trpos \vZ; \; \vW(\mu) \in \RR^{(Nm)\times N} \} $.  
Analogous to \feq{eq:GEP-CEX}, we set
\begin{equation}  \label{eq:GEP-N-CEX}
   \vhat{r}_N(\vmu) = \Ex_{\text{GPE}}(\vr_N(\vmu) | \vZ) = \arg 
      \min_{\vek{\chi}_N \in  \Svk_{N,\text{GPE}}(\vZ)} \Psi_{\vr_N}(\vek{\chi}_N)
      \in \Svk_{N,\text{GPE}}(\vZ)  
\end{equation}
with $\Psi_{\vr_N}(\vek{\chi}_N) = \Ex(\nd{\vr_N(\vmu) - \vek{\chi}_N(\vmu)}_{\Uvk}^2)
= \sum_{n=1}^N \var(r_N^n(\vmu) - \chi_N^n(\vmu))$.

The minimising weight matrix $\vW(\vmu)$ is 
\begin{equation}   \label{eq:Kriging-W-N}
  \vW(\vmu) = \tK_{N,\text{GPE}}^{-1} \vG(\vmu)\quad \text{ and } \quad \vhat{r}_N(\vmu) =
      \vG(\vmu)^\trpos \tK_{N,\text{GPE}}^{-1} \vZ;
\end{equation}
where $\tK_{N,\text{GPE}} = (\vK_N(\vmu_i,\vmu_j))_{i,j=1}^m$ and 
$\vG(\vmu) = [\vK_N(\vmu_1,\vmu),\dots,\vK_N(\vmu_m,\vmu)]^\trpos$.
Compare this to \feq{eq:CEX-1}, whence it becomes clear that \feq{eq:Kriging-W-N} is a 
special case of \feq{eq:CEX-1}.  This also indicates how to simply extend \feq{eq:Kriging-W-N}
for the case that the process $\vz$ may have a non-zero mean $\vbar{r}_N(\vmu)$.
Again, the Bayesian interpretation is that the prior knowledge is the Gaussian process $\vr_N(\vmu)$
with \emph{known} covariance matrix $\vK_N(\vmu_1,\vmu_2)$, as well as the mathematical form
of the manifold $\Svk_{N,\text{GPE}}(\vZ)$ which is used in the computation of the 
CEX in \feq{eq:GEP-N-CEX}.  For a recent error analysis of such GPE methods for parametric
problems using kernel methods, see \citem{BatlleppStuart2023}.  While it is conceivable
--- with the connection to conditional expectation as elaborated in \fsec{SS:re-inter-CEX} ---
to extend the Kriging sketched here in the direction of the filters used in the
approximation of the Bayesian update as described in \fsec{SS:filter-CEX}, to the author's
knowledge this has not yet been done.

The \emph{Reduced Basis Method} (RBM) (e.g.\ \citem{Quarteroni2015, HesthavenRozza2016})
is another example which may be given the interpretation of a conditional expectation.
In this context, compare \feq{eq:gen-CEX} with Propositions 3.1 and 3.3 in \citem{Quarteroni2015}.  
Let us remark here also that the computational efficiency of the RBM is often tied
to an affine dependence of the operator on the parameters $p\in \Pst$.
In \fsec{S:lin-maps} and the following developments, it was shown that any such
parametric problem can be \emph{re-parametrised}, so that in the new parameters
--- $\vmu \in \RR^N$ in the truncated and discretised form --- the problem is
indeed \emph{linear resp.\ affine}.

While an assumed connection with Gaussian processes is used in the procedures sketched above in
order to construct the ROM for new values of the parameters $\vmu\in \RR^N$, it had been noted
\citem{Neal1995} that broad \emph{artificial neural networks} --- such networks, 
as already mentioned several times, are an important and increasingly used class
of functions to evaluate the ROM for new values of the parameters $\vmu\in \RR^N$
\citem{Murphy2012, schmidh2014, Strang2019, Schwab2019, RaissiKarniadakis2019, KutyniokEtal2019,
FrescaEtal2020, Fleuret2023, NelsenStuart2024} --- at initialisation behave like
Gaussian processes.  This has become known as the neural network---Gaussian process correspondence
\citem{YangSalman2020, Yang-1-2021}, and is used in conjunction with kernel methods
\citem{BatlleppStuart2023} in the analysis of deep artificial neural networks.

\subsection{Parametric ROM as a random variable}  \label{SS:ROM-by-mini}
While in the previous \fsec{SS:ROM-RF} the parameters $\vmu\in \RR^N$ were interpreted
as elements of a set with which random vectors were parametrised --- a Gaussian random
field resp.\ process --- it is possible to take a different interpretation of a parametric
ROM as regards the parameters $\vmu\in \RR^N$.  To this end we continue with the assumptions 
made at the beginning of \fsec{SS:ROM-RF}, namely that the inner product on the space $\Mlg$
of functions of $\vmu \in \RR^N$ is given by a measure $\rho$: $\bkt{\vphi}{\psi}_{\Mlg} := 
\int_{\RR^N} \vphi(\vmu)\psi(\vmu)\, \rho(\di \vmu)$; and additionally 
we assume that $\rho$ is normalised 
($\rho(\RR^N) = 1$), so that it can be regarded as a probability measure.
Now $(\RR^N, \rho)$ is a probability space, so $\vr(\vmu)$ can be regarded as a 
$\RR^N$-valued random variable.  Going back to the last paragraphs in \fsec{SS:ROM-UQ},
there the additional uncertainties modelled on $\Omega_M \times \Omega_N$ were defined.
Thus, in total, one now has a $\RR^N$-valued random variable $\vr_MN(\vmu,\omega_M,\omega_N)$
on the total probability space $\Omega_t=\RR^N \times \Omega_M \times \Omega_N$ as 
already described there.  Additionally, there is a new expectation operator $\Ex_t$, which is an 
integration over this total probability space.

This means that all the developments described in \fsec{S:lin-maps} can now be applied here,
with $\C{U} \cong \RR^N$ and a new $\Pst = \RR^N \times \Omega_M \times \Omega_N$ which is
a probability space, and as $\C{Q} = \Lp_2(\Pst)$ one takes the $\Lp_2$-Hilbert space defined
by the total probability measure.  This means that one has the descriptions as detailed in
\fsecs{SS:correlation}{SS:ROMs} with the mappings $\vR_N$, $\vC_{\C{U},\mu}$, and $\vC_{\C{Q},\mu}$.  
And one may use the SVD methods there to find good approximations.  See 
\feq{eq:infin-dim-sing-numb} in Appendix~\ref{S:sing-val} for the definition of singular values
resulting from low-rank approximations, and \feq{eq:lin-rep} in \fsec{SS:correlation}
to arrive at the SVD of $\vR_N^\trpos = \sum_{j=1}^N \vsigma_j \, \vv_j \otimes s_j$.
It is well known that the truncated SVD is the approximation to this expression by
limiting the rank of the approximate mapping.  Written as a CEX, this becomes
\begin{equation}  \label{eq:SVD-N-CEX}
   \vhat{R}_{n,N} = \sum_{j=1}^n \vsigma_j \, \vv_j \otimes s_j 
   = \Ex_{t,F}(\vR_N | \vZ_n) = \arg \min_{\vX \in \Svk_{n,\text{SVD}}(\vZ_n)}
    \Psi_{\vR_N}(\vX)  \in \Svk_{n,\text{SVD}}(\vZ_n),
\end{equation}
where $\vZ_n = [\vv_1,\dots,\vv_n]$.  The manifold over which to minimise is 
\[
\Svk_{n,\text{SVD}}(\vZ_n) = \{ \vX \in \E{L}(\C{Q},\RR^N) \mid 
\vX = \vZ_n  \vek{\xi}_n = \sum_{j=1}^n \vv_j \otimes \xi_j, \;
  \vek{\xi}_n = [\xi_1,\dots,\xi_n]^\trpos \in \C{Q}^n \},
\]
and $\Psi_{\vR_N}(\vX) = \nd{\vR_N - \vX}^2_{\E{L}}$.  The novelty now is that in
$\Psi_{\vR_N}$ the mapping norm $\nd{\cdot}_{\E{L}}$ is used, which is not a Hilbert norm.
Therefore $\Ex_{t,F}$ has to be seen as a more general \emph{Fréchet mean}, and also
the minimisation is not like for a normal CEX over a subspace as in \fsec{SS:Var-CEX}, as
the manifold $\Svk_{n,\text{SVD}}(\vZ_n)$ --- these are special rank-$n$ maps --- is not a 
subspace; so that \feq{eq:SVD-N-CEX} may be classed as a Fréchet CEX.

However, for the ROM approximation $\vhat{r}_n$ of $\vr_N$, 
this is the best $n$-term approximation in $\C{U} \cong \RR^N$,
also known as the \KL{} expansion (KLE):
\begin{equation}  \label{eq:KLE-N-CEX}
   \vhat{r}_n = \Ex_{t}(\vr_N | \C{Z}_n) = \arg \min_{\vek{\chi} \in  \Svk_{n,\text{KLE}}(\C{Z}_n)}
    \Psi_{\vr_N}(\vek{\chi})   \in \Svk_{n,\text{KLE}}(\C{Z}_n)  
\end{equation}
where $\C{Z}_n = \{\vv_1,\dots,\vv_n\}$.  The manifold over which to minimise now is 
\[
\Svk_{n,\text{KLE}}(\vZ_n) = \{ \vek{\chi} \in \RR^N \mid 
   \vek{\chi} =  \sum_{j=1}^n \xi_j \vv_j , \;
  \xi_j \in \C{Q} \} = \spn \{\vv_1,\dots,\vv_n\},
\]
and $\Psi_{\vr_N}(\vek{\chi}) = \Ex_t(\nd{\vr_N - \vek{\chi}}^2_{\Uvk})$; 
so in total one has again a standard CEX.

The closely related POD technique uses $n < N$ samples resp.\ \emph{snapshots} grouped as
$\vV = \vZ_n = [\vv(\vmu_1),\dots,\vv(\vmu_n)]$ to construct a ROM basis.  In case  one 
desires a $k$-dimensional ($k < n$) ROM basis from these $n$ samples, that
approximation (\citem{chinestaWillcox2017}, Eq.(4)) may be approached 
similarly as a CEX with a manifold as follows: 
\[
\Svk_{n,\text{POD}}(\vZ_n) = \{ \vA \in \RR^{N \times n} \mid 
   \vA =  \vV \vV^\trpos \vZ_n, \;
  \vV \in \E{V}_k(\RR^N) \} ,
\]
where $\E{V}_k(\RR^N)$ is the Stiefel manifold of orthogonal $N \times k$ matrices
and $\Psi_{\vZ_N}(\vA) = \Ex_t(\nd{\vZ_n - \vA}^2_{F})$ 
($\nd{\cdot}_F$ is the Frobenius norm).  The POD basis is again given as a CEX:
\begin{equation}  \label{eq:POD-N-CEX}
   \vV_{k,\text{POD}} = \Ex_{t}(\vV | \vZ_n) = \arg \min_{\vA \in  
      \Svk_{n,\text{POD}}(\vZ_n)}
    \Psi_{\vZ_N}(\vA)  \in  \Svk_{n,\text{POD}}(\vZ_n)  
\end{equation}

The reduced basis method (RBM) \citem{chinestaWillcox2017} (cf.\ also the monographs
\citem{Quarteroni2015, HesthavenRozza2016}, as well as \citem{BennWilcox-paramROM2015} and 
the work described in e.g.\ \citem{DrohHass2012, Quarteroni2014, CohenEtal2015,  MoRePaS2015})
is for our purposes best considered in the context of solving
a linear elliptic coercive parametric PDE, i.e.\ finding the solution $u(\vmu) \in \Uvk$ such that
\begin{equation}  \label{eq:RBM-GE}
 a(\vmu,u(\vmu),v) =\ip{A(\vmu) u(\vmu)}{v}_{\Uvk} = F(v) = \ip{f}{v}_{\Uvk} \quad 
   \forall v \in \Uvk_{\text{RBM}},
\end{equation}
where $A$ is the self-adjoint operator $A:\Uvk \to \Uvk^*$ associated to the bilinear 
form $a(\vmu;\cdot,\cdot)$.  It is well know that this is equivalent to minimising
the functional $E(v)=\frk{1}{2}\ip{A(\vmu) v}{v}_{\Uvk} - \ip{f}{v}_{\Uvk}$, but also the
quadratic form $Q_{\vmu}(v) = \frk{1}{2}\ip{A(\vmu)(v - u(\vmu))}{(v - u(\vmu)}_{\Uvk}$, where
symbolically $u(\vmu) = A^{-1}(\vmu)f$.
Having computed the snapshots $\C{Z}_n = \{u(\vmu_1),\dots,u(\vmu_n)\}$, the RBM solution
for a new value $\vmu$ is computed by minimising $E(v)$ over 
$\Svk_{n,\text{RBM}}(\C{Z}_n) = \spn \, \C{Z}_n$, which is the Galerkin solution of \feq{eq:RBM-GE}
with $\Uvk_{\text{RBM}} = \Svk_{n,\text{RBM}}(\C{Z}_n)$.  Taking into account the other 
modelling and numerical uncertainties described before, define the quadratic functional 
$\Psi_{u(\vmu),\text{RBM}}(v) = \Ex_g(Q_{\vmu}(v))$, which is a generalised and extended \
least-squares expression.  Then, an extended RBM solution is given by the CEX
\begin{equation}  \label{eq:RBM-N-CEX}
   u_{n,\text{RBM}}(\vmu) = \Ex_{t}(u(\vmu) | \C{Z}_n) = \arg \min_{v \in \Svk_{n,\text{RBM}}(\C{Z}_n)}
    \Psi_{u(\vmu),\text{RBM}}(v)  \in \Svk_{n,\text{RBM}}(\C{Z}_n).  
\end{equation}


As a last example, we would like to mention the low-rank tensor computation of a ROM;
for general material related to low-rank tensor representations, cf.\ \citem{Hackbusch_tensor}.
As seen previously, a parametric solution to \feq{eq:RBM-GE} is a function
$u(\vmu,\omega_M,\omega_N)$, where the short-hand notation of the uncertainties coded with
$ (\omega_M,\omega_N))$ has been kept for the sake of brevity.  With the repeated application
of methods from \fsec{S:lin-maps}, one can see that this can be written as a tensor product
$u(\vmu,\omega_M,\omega_N) = \sum_j v_j(\vmu) s_{j,M}(\omega_M) s_{j,N}(\omega_N)$, 
with $v_j(\vmu) \in \Uvk$ and RVs $s_{j,M}(\omega_M)$ and $s_{j,N}(\omega_N)$.  
One way of approximating this kind of solution \citem{falco-nouy-2011} is to define the 
nested sequence of sub-manifolds $\Svk_k = \{ w \mid w = \sum_{j=1}^k v_j(\vmu) s_{j,M}(\omega_M) 
s_{j,N}(\omega_N)\}$ for $k \in \D{N}$.  

From the functional $\Psi_{u(\vmu),\text{RBM}}(v)$ from \feq{eq:RBM-N-CEX}, let us define the new
functional $\Psi_{k}$ on $\Svk_k$.  Assuming that the minimisation of $\Psi_{k-1}$ on $\Svk_{k-1}$
has already been carried out and
given as result the rank-$(k-1)$ approximation $u_{k-1}(\vmu)$, we now define 
\[
\Psi_{k}(v_k(\vmu) s_{k,M}(\omega_M) s_{k,N}(\omega_N)) = 
   \Psi_{u_{k-1}(\vmu),\text{RBM}}(w_k(\vmu,\omega_M,\omega_N)),
\]
where $w_k(\vmu,\omega_M,\omega_N) := u_{k-1}(\vmu) + v_k(\vmu) s_{k,M}(\omega_M) s_{k,N}(\omega_N)$,
on $\Svk_k$.  Then, the rank-$k$ approximation is computed as
\begin{equation}  \label{eq:Tens-N-CEX}
   u_{k}(\vmu) = \Ex_{t}(u(\vmu) | u_{k-1}(\vmu)) = \arg \min_{v_k s_{k,M} s_{k,N} \in \Svk_1(\C{Z}_n)}
    \Psi_{u_{k-1}(\vmu)}(v_k s_{k,M} s_{k,N})  \in \Svk_k(\C{Z}).  
\end{equation}
The minimisation in \feq{eq:Tens-N-CEX} is often computed as an \emph{alternating minimisation}.
Let us also point out that there are other ways of defining possible
low-rank tensors, cf.\ \citem{Hackbusch_tensor}.

\section{Conclusions}  \label{S:concl}
Here the main goal was to try and formulate a few ways of computing parametric 
reduced order models (ROMs) as a conditional
expectation (CEX).  To this end in \fsec{S:lin-maps} a connection has been made between parametric
elements and their encoding in linear maps.  This led, with the help of spectral theory (summarised
in Appendix~\ref{S:spec-dec}) and the singular value decomposition (SVD) (summarised
in Appendix~\ref{S:sing-val}), to various other decompositions and expansion like the \KL{}
expansion (KLE).  The variational theory of expectation and conditional expectation was sketched in
\fsec{S:CEX}.

Connecting everything in \fsec{S:ROM-CEX}, we first analysed all the uncertainties inherent
in establishing a ROM, and then showed that many different ROM computations could  be
interpreted as a CEX computation.  Specifically, attention was focused on Gaussian process
emulation (GPE) and kernel methods, as well as the reduced basis methods (RBM), the truncated
SVD and the proper orthogonal decomposition (POD), and, last but not least, low-rank
tensor approximations.

Although the parameter set $\C{P}$ was not assumed to have any inherent structure, one
assumption was that the assignment $\C{P} \ni p \to r(p) \in \Uvk$ was unique, i.e.\ was
a mapping.  Other than this, no other assumptions were made regarding either the
parameter set $\C{P}$ nor the kind of mapping.  Obviously, if such additional assumptions
are made, they can be exploited in the ROM calculations.  In a Bayesian spirit, this
may be seen as additional a priori information.

On the other hand, in future work a weakening of the assumption of a mapping 
$\C{P} \ni p \to r(p) \in \Uvk$ may be explored.  Such weaker assumptions would allow one
to be able to approach problems which involve bifurcations and multiple solutions for
a single parameter value.




%
\subsection*{Abbreviations}
\noindent 
\begin{tabular}{@{}ll}
RKHS & Reproducing Kernel Hilbert Space\\
CONS & Complete Ortho-Normal System\\
EnKF & Ensemble Kalman filter\\
GMKF & Gauss-Markov Kalman filter\\
PCKF & Polynomial Chaos Kalman filter\\
PDE & Partial Differential Equation\\
QoI & Quantity of Interest\\
KLE & \KL{} Expansion\\
FOM & Full Order Model\\
ROM & Reduced Order Model\\
SVD & Singular Value Decomposition\\
POD & Proper Orthogonal Decomposition\\
PGD & Proper Generalised Decomposition\\
RBM & Reduced Basis Method\\
pdf & probability density function\\
CEX & Conditional Expectation\\
UQ  & uncertainty quantification\\
KF  & Kalman filter\\
BU  & Bayesian updating\\
MC  & Monte Carlo\\
ML  & machine learning\\
RV  & Random Variable
\end{tabular}

\subsection*{Acknowledgments}
{The research reported in this publication was partly supported by  
   a Gay-Lussac Humboldt Research Award.}




\bibliographystyle{hgmplain-1}
\bibliography{./bib/jabbrevlong,./bib/mat_BU-1-S,./bib/phys_D,./bib/fa,./bib/risk,%
./bib/stochastics,./bib/fuq-new,./bib/sfem,./bib/highdim,./bib/filtering,./bib/inverse,./bib/new}

\appendix
\section[\appendixname~\thesection]{Spectral Decomposition}  \label{S:spec-dec}
{First a reminder} of the {finite dimensional} theory:  Let $\vek{A} = \vek{A}^\trpos$
be a self-adjoint $n \times n$-matrix.  Then the spectrum $\sigma(\vek{A}) \subset \RR$ is real, 
and further it is assumed that the eigenvalues $\lambda_k \in \sigma(\vek{A})$
are ordered as a subset of $\RR$, i.e.\ $\lambda_1 \le \dots \lambda_k \le \dots \le \lambda_n$
with corresponding CONS of eigenvectors $\vek{u}_k$.  
Observe that $\vek{a} \otimes \vek{b} = \vek{a} \vek{b}^\trpos$, so that the
projection on the eigen-space of $\lambda_m$ is $\vek{\Pi}_m = \sum_{\lambda_k = \lambda_m} 
\vek{u}_k \otimes \vek{u}_k =: \upDelta \vek{E}_{m}$; and thus   
$\vek{I} = \sum_{m=1}^\ell \vek{\Pi}_m$ (where $\ell$ is the number of distinct eigenvalues)
is a partition of unity.  
As a side remark, the projection $\vek{\Pi}_m$ is in fact a polynomial in $\vek{A}$, 
easily seen to equal the elementary Lagrange polynomial $\vek{L}_m(\vek{A}) = 
\prod_{\lambda_k \ne \lambda_m} \frac{1}{\lambda_m - \lambda_k}
  (\vek{A} - \lambda_k \vek{I}) = \vek{\Pi}_m$.
Then there are three common ways of writing the spectral decomposition:
 
\begin{description}
\item[Sum of projections:] Set $\vek{E}_0 = \vek{0}$, and for the orthogonal spectral projections
  onto the eigen-spaces with $\lambda_k \le \lambda_m$:
  $\vek{E}_m = \vek{E}_{m-1} + \upDelta \vek{E}_{m} = \vek{E}_{m-1} + \vek{\Pi}_m$ 
  ($1 \le m \le \ell; \vek{E}_\ell = \vek{I}$).  
  This can also be written as $\vek{E}_m = \bbbone_{]-\infty, \lambda_m]}(\vek{A}) = 
  \sum_{k \le m} \vek{L}_k(\vek{A})$, it is a polynomial in $\vek{A}$.  
  Then the spectral decomposition of $\vek{A}$ can be
  written as a sum of orthogonal projections
  \begin{equation}    \label{eq:fin-dim-sum-proj}
  {\vek{A} = \sum_{m=1}^\ell \lambda_m \, \upDelta \vek{E}_m} = 
  \sum_{m=1}^\ell \lambda_m \, \vek{\Pi}_m .
  \end{equation}
\item[Diagonalisation:] Collect the eigenvectors $\vek{u}_k$ as the $k$-th column into
  the orthogonal resp.\ unitary matrix $\vek{U}$, and the eigenvalues into the
  diagonal matrix $\vek{\Lambda} = \diag(\lambda_k)$.  Then $\vek{A}$ is unitarily
  equivalent to multiplication by a diagonal matrix:
  \begin{equation}    \label{eq:fin-dim-diag}
  \vek{A} = \vek{U} \vek{\Lambda} \vek{U}^\trpos .
  \end{equation}
\item[Tensor decomposition:]  This is simply a component-wise expansion of \feq{eq:fin-dim-diag}: 
  \begin{equation}    \label{eq:fin-dim-tens}
  \vek{A} = \sum_{k=1}^n \lambda_k \, \vek{u}_{k} \otimes \vek{u}_{k} .
  \end{equation}
\end{description}

Now let $\vek{R}$ be a $m \times n$-{matrix} 
with $\ker \vek{R} = \{ 0 \}$ (this implies $m \ge n$), 
$\vek{C}_{\Uvk}=\vek{R}^\trpos \vek{R} \in \RR^{n \times n}$, and 
  $\vek{C}_{\C{Q}} = \vek{R} \vek{R}^\trpos  \in \RR^{m \times m}$.
Then their {spectral-} and {singular value}-decomposition (SVD) as diagonalisation 
follows from \feq{eq:fin-dim-diag}:
\begin{equation}    \label{eq:fin-dim-SD-SVD-diag}
  \vek{C}_{\Uvk} = \vek{V} \vek{\Lambda}_{\Uvk} \vek{V}^\trpos, \quad 
  \vek{C}_{\C{Q}} = \vek{S} \vek{\Lambda}_{\C{Q}} \vek{S}^\trpos, \quad 
  \vek{R} = \vek{S} \vek{\Sigma} \vek{V}^\trpos ,
\end{equation}
where $\vek{V} \in \RR^{n \times n}$ and $\vek{S}  \in \RR^{m \times m}$ are 
orthogonal resp.\ unitary, $\vek{\Sigma}  \in \RR^{m \times n}$
is a diagonal {matrix} of singular values, and $\vek{\Lambda}_{\Uvk} =
\vek{\Sigma}^\trpos \vek{\Sigma} \in \RR^{n \times n}$, 
$\vek{\Lambda}_{\C{Q}} = \vek{\Sigma} \vek{\Sigma}^\trpos \in \RR^{m \times m}$
are the diagonal matrices of eigenvalues $\lambda_k = \vsigma^2_k$, 
with $\vsigma_k$ ($ 1 \le k \le n$) the $k$-th singular value.  
These diagonalisations immediately suggest other factorisations
of $\vek{C}_{\Uvk}$ (or $\vek{C}_{\C{Q}}$) and resulting representations of $r(p)$, namely
\begin{equation}    \label{eq:fin-dim-SD-fact}
  \vek{C}_{\Uvk} = (\vek{V} \vek{\Lambda}_{\Uvk}^{\frk{1}{2}}) 
  (\vek{V} \vek{\Lambda}_{\Uvk}^{\frk{1}{2}})^\trpos , \quad  
  \vek{C}_{\Uvk} = \vek{C}_{\Uvk}^{\frk{1}{2}} \vek{C}_{\Uvk}^{\frk{1}{2}}, \quad \text{ where }\;
  \vek{C}_{\Uvk}^{\frk{1}{2}} = \vek{V} \vek{\Lambda}_{\Uvk}^{\frk{1}{2}} \vek{V}^\trpos.     
\end{equation}

The variant of \feq{eq:fin-dim-tens} based on \feq{eq:fin-dim-SD-SVD-diag} is
\begin{equation}    \label{eq:fin-dim-SD-SVD-tens}
  \vek{C}_{\Uvk} = \sum_{k=1}^n \vsigma^2_k \, \vek{v}_k \otimes \vek{v}_k, \quad
  \vek{C}_{\C{Q}} = \sum_{k=1}^n \vsigma^2_k \, \vek{s}_k \otimes \vek{s}_k, \quad
  \vek{R} = \sum_{k=1}^n \vsigma_k \, \vek{s}_k \otimes \vek{v}_k.
\end{equation}

The decomposition according to \feq{eq:fin-dim-sum-proj} is rarely used in the
finite-dimensional theory.  But in the infinite-dimensional case \citem{dautrayLions-03}, 
when $A$ is assumed to be a self-adjoint linear operator on a Hilbert space $\C{H}$, where
the spectrum may have continuous parts, it
is the most commonly known form of the spectral decomposition:

\begin{description}
\item[Spectral measure:] On the (Borel) space $(\D{R},\F{B})$ there is a measure
   $E:\F{B} \to \E{L}(\C{H})$ whose values are commuting orthogonal projections ---
   a projection valued measure --- with 
   $E(\emptyset) = 0,\; E(\D{R}) = I,\;  E(]a, b]) = \bbbone_{]a, b]}(A)$, such that
   --- corresponding to \feq{eq:fin-dim-sum-proj} ---
  \begin{equation}    \label{eq:proj-val-meas}
    {A = \int_{\sigma(A)} \lambda\, E(\di \lambda)} .
  \end{equation}
  In case of a purely discrete spectrum with the distinct eigenvalues $\lambda_k \in 
  \sigma(A)$, and with $\upDelta E_k := \bbbone_{]\lambda_{k-1}, \lambda_k]}(A)$, one has
  the direct equivalent of \feq{eq:fin-dim-sum-proj}, the discrete version of \feq{eq:proj-val-meas}:
  \begin{equation}    \label{eq:infin-dim-sum-proj}
    A = \sum_{\lambda_k\in\sigma(A)} \lambda_k \upDelta E_k  .
  \end{equation}
\item[Diagonalisation:] Less known than \feq{eq:proj-val-meas} but still fairly standard is:
  the self adjoint $A\in \E{L}(\C{H})$ is unitarily equivalent to a multiplication operator 
  {$M_k \in \E{L}(\Lp_2(\C{T}))$}  by a real-valued function $k(t)$ on some 
  $\Lp_2(\C{T}) \ni \xi(t) \mapsto (M_k \xi)(t) = k(t)\xi(t) \in \Lp_2(\C{T})$,
  where $\C{T}$ is some measure space.
  The spectrum $\sigma(A)$ is the (essential) range of $k:\C{T}\to\RR$.  This is the
  direct analogue of \feq{eq:fin-dim-diag}:
  \begin{equation}    \label{eq:infin-dim-diag}
     A = U M_k U^\dagger ,
  \end{equation}
  where $U\in\E{L}(\Lp_2(\C{T}),\C{H})$ is unitary, $U^\dagger$ is its adjoint, and the
  multiplication operator $M_k$ may be seen as a ``diagonal'' operator.  
\item[Nuclear Spectral Theorem:] In case $A \in \E{L}(\C{H})$ has a partly continuous
  spectrum, a nuclear Gel'fand triple $\Phi \hookrightarrow \C{H} \hookrightarrow \Phi^*$ 
  (a ``rigged'' Hilbert space) with $A\Phi \subseteq \Phi$ and generalised
  eigenvectors in $\Phi^*$ are needed, e.g.\ \citem{dautrayLions-03}.  
  Here we restrict ourselves for the sake of brevity and simplicity
  to the discrete case where each spectral value 
  $\lambda_k \in \sigma(A)$ is an eigenvalue, counted according to multiplicity,
  with orthonormal eigenvectors $u_k \in \C{H}$.  Then the analogue of \feq{eq:fin-dim-tens} is:
  \begin{equation}    \label{eq:infin-dim-tens}
     A = \sum_{k} \lambda_k\; u_k \otimes u_k.
  \end{equation}
\end{description}

Let $R: \Uvk \to \C{Q}$ be an injective linear map as in \fsec{SS:correlation}.
The statement \feq{eq:proj-val-meas} is typically not used to factor $R$
via $C_{\Uvk} = R^\dagger R \in \E{L}(\Uvk)$, see \fsec{SS:correlation}.
But the statement \feq{eq:infin-dim-diag} is.
So one has, together with $C_{\C{Q}} = R R^\dagger  \in \E{L}(\C{Q})$, a
{spectral-} and {singular value}-decomposition (SVD) as analogue of
the diagonalisation \feq{eq:fin-dim-SD-SVD-diag}:
\begin{equation}    \label{eq:infin-dim-SD-SVD-diag}
  C_{\Uvk} = V M_{\vsigma}^2 V^\dagger, \quad 
  C_{\C{Q}} = S M_{\vsigma}^2 S^\dagger, \quad 
  R = S M_{\vsigma}  V^\dagger , \quad 
  R^\dagger = V M_{\vsigma}  S^\dagger,
\end{equation}
where $V \in \E{L}(\Lp_2(S), \Uvk)$ and $S \in \E{L}(\Lp_2(\C{T}), \C{Q})$
are unitary, and $M_{\vsigma} \in \E{L}(\Lp_2(\C{T}))$ is a multiplikation
operator with the positive function $\vsigma(t)$.  The essential range of
$\vsigma$ may be seen as a possible continuous generalisation of the singular 
values, which is possible due to the fact that $R$ was assumed injective,
so that $S$ is unitary instead of just a partial isometry.

Now define $s \in \C{Q}$ via $s(p) := S M_{\vsigma}^{-1} V^\dagger r(p)$.  
Then the continuous SVD of $R^\dagger$ leads to the following representation
of $r(p)$, linear in $s(p)$:
\begin{equation}    \label{eq:infin-dim-SD-fact-rep} 
    r(p) =  R^\dagger s(p) = V M_{\vsigma}  S^\dagger s(p) .
\end{equation}
To use this \feq{eq:infin-dim-SD-fact-rep} to define a ROM, set on $\Lp_2(\C{T})$
for $0 < a < \nd{\vsigma}_\infty$ as a cut-off:
$\zeta_a(t) = \bbbone_{\{\vsigma \ge a\}}(t) \vsigma(t)$.  Then the approximation
\begin{equation}    \label{eq:infin-dim-SD-fact-ROM} 
    r_a(p) =   V M_{\zeta_a}  S^\dagger s(p) \approx r(p) .
\end{equation}
is a kind of continuous ROM.

Again, the diagonalisations \feq{eq:infin-dim-SD-SVD-diag} immediately suggest other 
factorisations, as in \feq{eq:fin-dim-SD-fact},
of $C_{\Uvk}$ (or $C_{\C{Q}}$) and resulting representations of $r(p)$, namely
\begin{equation}    \label{eq:infin-dim-SD-fact}
  {C}_{\Uvk} = ({V} M_{\vsigma}) ({V} M_{\vsigma})^\dagger , \quad  
  {C}_{\Uvk} = {C}_{\Uvk}^{\frk{1}{2}} {C}_{\Uvk}^{\frk{1}{2}}, \quad \text{ where }\;
  {C}_{\Uvk}^{\frk{1}{2}} = {V} M_{\vsigma} {V}^\dagger .     
\end{equation}

In case $C_{\Uvk}$ (and $C_{\C{Q}}$) has a purely discrete spectrum with
eigenvalues $\sigma_m^2$ and eigenvectors $v_m$ (and eigenfunctions $s_m$ of $C_{\C{Q}}$),
the variant of \feq{eq:fin-dim-SD-SVD-tens} based on \feq{eq:infin-dim-tens} is
\begin{equation}    \label{eq:infin-dim-SD-SVD-tens}
  {C}_{\Uvk} = \sum_{m} \vsigma^2_m \, {v}_m \otimes {v}_m, \quad
  {C}_{\C{Q}} = \sum_{m} \vsigma^2_m \, {s}_m \otimes {s}_m, \quad
  {R} = \sum_{m} \vsigma_m \, {s}_m \otimes {v}_m .
\end{equation}

\section[\appendixname~\thesection]{Singular Values and Singular Numbers}  \label{S:sing-val}
In \feq{eq:infin-dim-SD-SVD-diag} a kind of continuous SVD of $R$ on infinite-dimensional
Hilbert spaces was possible due to the special nature of $R$.  In the finite-dimensional
setting one is reduced to \feq{eq:fin-dim-SD-SVD-tens}, or even in the purely discrete
case to \feq{eq:infin-dim-SD-SVD-tens},  and those singular values coincide with the
\emph{singular numbers} (s-numbers) to be defined below, e.g.\ \citem{KowalskiEtal1995}. 

These \emph{singular numbers} (s-numbers) are defined via low-rank approximations, another 
possibility to produce a ROM.
Let $A: \C{H}_1 \to \C{H}_2$ be a bounded linear operator between Hilbert spaces $\C{H}_1, \C{H}_2$,
with the {operator norm} is defined as
$\nd{A}_{1,2} = \sup_{0 \ne x \in \C{H}_1} {\nd{Ax}_2}/{\nd{x}_1}$.

Recalling that the
{rank} of a linear map $B: \C{H}_1 \to \C{H}_2$ is $\rank B = \dim(\im B$),
one defines
the {singular numbers} (s-numbers) $\{\zeta_k(A)\}_{k=1,\dots}$ of such a
$A \in \E{L}(\C{H}_1, \C{H}_2)$ as: 
\begin{equation}    \label{eq:infin-dim-sing-numb}
   \zeta_k(A) := \inf \{\nd{A - B}_{1,2} \mid B \in \E{L}(\C{H}_1, \C{H}_2), \quad \rank B < k \},
   \quad k=1,\dots .
\end{equation}
This implies
\begin{equation}  \label{eq:infin-dim-sing-numb-k}
   \nd{A}_{1,2} = \zeta_1(A) \ge \zeta_2(A) \ge \dots \ge 0, \quad 
   \text{ and } \; \zeta_k(A) = \zeta_k(A^\dagger) ,
\end{equation}
as well as the following:
If  $A$ is compact, then $\zeta_k(A) \to 0$, and $\zeta_k(A)^2$ is an eigenvalue of
   $A^\dagger A$  and $A A^\dagger$.  Additionally, in each of the two Hilbert spaces
   there exists a CONS $\{e^\ell_k\}\subset \C{H}_\ell$ such that  
\[   
   \zeta_k(A) = \nd{A - \sum_{j=1}^{k-1}\zeta_j(A) e^2_j \otimes e^1_j}_{1,2},\quad \text{ and}
\]
\[
   A e^1_k = \zeta_k(A) e^2_k; \quad A^\dagger e^2_k = \zeta_k(A) e^1_k.
\]
Obviously the CONS $\{e^1_k\}\subset \C{H}_1$ is an eigenvector basis for $A^\dagger A$, and
the CONS $\{e^2_k\}\subset \C{H}_2$ is an eigenvector basis for $A A^\dagger$.
This means that for compact operators the singular numbers (s-numbers) and singular values coincide.

Let us mention that the  (s-numbers) are connected with the so-called n-widths in
approximation theory, see e.g.\ \citem{KowalskiEtal1995, FloaterEtal2021};
an important topic which for the sake of brevity will not be pursued further here.

\end{document}